\documentclass[acmsmall,compsoc]{acmart}
\usepackage{xcolor}
\usepackage[T1]{fontenc}
\usepackage{amsmath}
\usepackage{enumitem}
\usepackage{multirow}
\usepackage{wrapfig,lipsum,booktabs}
\usepackage[linesnumbered, lined, ruled]{algorithm2e}
\usepackage{mathrsfs}
\usepackage{graphicx,subfigure}
\usepackage{parallel,enumitem}

\theoremstyle{plain}

\newtheorem{assumption}{Assumption}[section]

\AtBeginDocument{%
  \providecommand\BibTeX{{%
    \normalfont B\kern-0.5em{\scshape i\kern-0.25em b}\kern-0.8em\TeX}}}

\setcopyright{none}
\acmJournal{TORS}
\acmVolume{0}
\acmNumber{0}
\acmArticle{0}
\acmMonth{0}

\begin{document}

\title{Label Denoising through Cross-Model Agreement}

\author{Yu Wang}
\affiliation{%
  \institution{University of Science and Technology of China}
  \country{China}
%   \city{Hefei}
%   \country{China}
}
\email{wy2001@mail.ustc.edu.cn}

\author{Xin Xin}
\authornote{Xin Xin is the corresponding author.}
\affiliation{%
  \institution{Shandong University}
  \country{China}
  }
\email{xinxin@sdu.edu.cn}

\author{Zaiqiao Meng}
\affiliation{%
  \institution{University of Glasgow}
  \country{UK}
}
\email{zaiqiao.meng@gmail.com}

\author{Joemon M Jose}
\affiliation{%
 \institution{University of Glasgow}
 \country{UK}
 }
\email{Joemon.Jose@glasgow.ac.uk}

\author{Fuli Feng}
\affiliation{%
 \institution{University of Science and Technology of China}
 \country{China}
 }
\email{fulifeng93@gmail.com}

\renewcommand{\shortauthors}{Trovato and Tobin, et al.}

\newcommand{\wy}[1]{{\color{black}{#1}}}
\newcommand{\wyy}[1]{{\color{black}{#1}}}
\newcommand{\xx}[1]{{\color{black}{#1}}}
\newcommand{\new}[1]{{\color{black}{#1}}}
\newcommand{\neww}[1]{{\color{black}{#1}}}

\begin{abstract}
Learning from corrupted labels is very common in real-world machine-learning applications. Memorizing such noisy labels could affect the learning of the model, leading to sub-optimal performances. In this work, we propose a novel framework to learn robust machine-learning models from noisy labels. Through an empirical study, we find that different models make relatively similar predictions on clean examples, while the predictions on noisy examples vary much more across different models. Motivated by this observation, we propose \em denoising with cross-model agreement \em (DeCA) which aims to minimize the KL-divergence between the true label distributions parameterized by two machine learning models while maximizing the likelihood of data observation. 

We employ the proposed DeCA on both the binary label scenario and the multiple label scenario. For the binary label scenario, we select implicit feedback recommendation as the downstream task and conduct experiments with four state-of-the-art recommendation models on four datasets. For the multiple-label scenario, the downstream application is image classification on two benchmark datasets. Experimental results demonstrate that the proposed methods significantly improve the model performance compared with normal training and other denoising methods on both binary and multiple-label scenarios.

%   In this paper, we propose probabilistic and variational denoising methods (PVD) for corrupted labels. 

%   Through an empirical study, we find that \wy{models trained with different random seeds} tend to make relatively similar predictions on clean labels while predictions on noisy labels vary much more across different models. Motivated by this observation, we propose to denoise labels by minimizing the KL-divergence between the real label distributions parameterized by two different models while maximizing the likelihood of data observation.
% \textcolor{red}{We then show that DPI recovers the  lower bound evidence of an variational auto-encoder when the real user preference is considered as the latent variables.} 
% We then show that this method recovers the evidence lower bound of an variational auto-encoder when the real label distribution is considered as the latent variables. We employ the proposed denoising method on two downstream tasks: implicit feedback recommendation and image classification, corresponding to binary labels and multi-class labels, respectively. Experimental results demonstrate that the proposed methods significantly improve the model performance compared with normal training and other denoising methods. Codes will be open-sourced.
\end{abstract}




\maketitle

\section{Introduction}
\label{introduction}

Supervised learning is widely used in real-world applications, including all kinds of fields like computer vision\cite{lei2016skin,li2014medical}, natural language processing\cite{minaee2021deep,dien2019article}, information filtering\cite{bobadilla2020deep,he2017neural} and bio-informatics\cite{tang2019recent, min2017deep}. Recently, deep learning models have achieved great success in both academic and industrial applications. However, \citet{understand} shows that the high model capability of deep learning models enable them to memorize not only the expected clean training samples but also the noisy data with corrupted or even random labels. While in real-world scenarios, learning from corrupted labels is extremely common. For example, to address the information overload problem, recommender systems are widely used in various online services such as e-commerce \cite{lin2019cross}, multimedia platforms \cite{davidson2010youtube,van2013deepspotify} and social media \cite{chen2019efficient}. 
A recommendation agent is usually trained using implicit feedback (e.g. view and  click behaviours) since implicit data is much easier to be collected compared with explicit ratings \cite{rendle2014bayesian,he2016fast}. However, the collected implicit feedback is corrupted by various kinds of biases\cite{chen2020bias}, such as the popularity bias\cite{abdollahpouri2019managing}, the exposure bias\cite{khenissi2020modeling,abdollahpouri2020multi}, etc. Such corrupted data can lead to misunderstanding of the real user preference and sub-optimal recommendation \cite{chen2020bias,wen2019leveraging}.
In the field of computer vision, collecting external expert labels is rather expensive and as a result, learning from corrupted crowd-sourcing labels is also the common use case \cite{shen2019learning,shu2019meta,nguyen2019self}. Generally speaking, effective denoising methods for learning from corrupted labels are of vital importance to deploy machine learning models in practical applications.
To learn from corrupted labels, some denoising methods have been done by using re-sampling methods \cite{yu2020sampler,gantner2012personalized,ding2019samplerview,ding2018improvedview} or re-weighting methods \cite{wang2020denoising,jiang2018mentornet,ren2018learning}.
Re-sampling methods focus on designing more effective data samplers. For example, \citet{gantner2012personalized} considers that for the training of a recommendation agent, the missing interactions of popular items are highly likely to be real and clean negative samples. However, the performance of re-sampling methods depends heavily on the sampling distribution and suffers from high variance \cite{yuan2018fbgd}. \citet{wang2020denoising} proposed a re-weighting method which assigns lower weights or zero weights to high-loss samples since the noisy examples would have higher losses. However, \citet{shu2019meta} shows that hard yet clean examples also tend to have high losses. As a result, such re-weighting methods could encounter difficulties to distinguish between hard clean and noisy examples. Some efforts utilize self-paced learning or curriculum learning to assign weights for training samples \cite{jiang2018mentornet,nguyen2019self,ren2018learning,shu2019meta,bengio2009curriculum}, while such kinds of methods usually need to involve a clean set for the training of the teacher model. Besides, some researches focus on utilizing auxiliary information to denoise corrupted labels \cite{lu2019effects,kim2014modeling,liu2010understanding,yi2014beyond} but these kinds of methods need additional knowledge input.

In this work, we propose a weighting-free \emph{\textbf{De}noising with \textbf{C}ross-Model \textbf{A}greement} (DeCA) method for learning from corrupted labels without using additional data knowledge. The signal for denoising comes from an insightful observation: different models tend to make relatively similar predictions for clean samples with correct labels, while predictions for noisy samples with corrupted labels would vary much more among different models. This observation is substantiated in an empirical study described in Section \ref{sec:motivation}.
To this end, we propose DeCA which aims to minimize the Kullback–Leibler (KL) divergence between the real label distributions parameterized by two different models, and meanwhile, maximize the likelihood of the data observation given the true label distribution. We then show that the proposed DeCA recovers the evidence lower bound (ELBO) of a variational auto-encoder (VAE) in which the true label acts as the latent variable. 
DeCA can be considered as a learning framework which incorporates the weakly supervised signals from different model predictions to denoise the target model.
The proposed DeCA can be naturally incorporated with existing machine learning models. To verify the effectiveness and generalization of DeCA, we conduct experiments on two downstream tasks: implicit feedback recommendation and image classification, corresponding to binary labels and multi-class labels, respectively. 
\wy{More precisely, we start with the binary classification problem in the field of implicit feedback recommendation. Then in order to extend our proposed framework to the multiple label scenario, we follow the settings in \cite{shu2019meta, shen2019learning, nguyen2019self} where they introduce the general framework for denoising and then apply their methods for image classification. To make a fair comparison, we also select image classification as the downstream task for the multiple-label scenario.}

\new{This manuscript presents an extension of our prior work, DeCA \cite{DeCA}, initially developed as a denoising framework for binary classification scenarios. The progression from binary to multi-class classification is seemingly straightforward but encompasses several intrinsic challenges. One complication stems from the complexity of noisy labels in the multi-class scenario. Unlike binary classification, where a noisy label can only be one alternative, in a multi-class environment, a noisy label could originate from any class within the label set. This characteristic may complicate the process of identifying correlations between these labels, which is an important aspect of denoising. Furthermore, the inherent structure of the DeCA framework poses another challenge for the multi-class extension. DeCA relies on iterative training and constructs the mapping from true labels to noisy labels using two distinct models. This methodology, while effective in a binary context, does not translate directly to multi-class scenarios. The increased complexity requires a more nuanced approach, necessitating the refinement of the architecture and training protocols. 
} \neww{While it's not inherently true that recommendation tasks are simpler than image classification tasks - as one might initially assume given the binary versus multi-class nature of these tasks - applying our method presents more significant challenges in the context of multi-class classification tasks.}

To summarize, the main contributions of this work are as follows:
\begin{itemize}[leftmargin=*]
  \item We find that the different models tend to make more similar predictions for clean examples than for noisy ones. This observation provides new denoising signals to devise effective learning methods for corrupted labels.
  \item We propose the learning framework DeCA which utilizes the difference between model predictions as the denoising signal for corrupted labels and infer the real label distribution from corrupted data observation. \new{Our framework is designed to be more unified compared to its earlier version presented in \cite{DeCA}, as it extends to multi-class settings while maintaining compatibility with binary classification scenarios. } 
  \item We instantiate DeCA with multiple state-of-the-art machine learning models on both binary implicit feedback data and multi-class image classification data.  Extensive experimental results on benchmark datasets demonstrate the effectiveness and generalization ability of our proposed methods.
\end{itemize}

\section{Background and Related Work} % (fold)
\label{sec:related_work}
% Implicit feedback has been widely used in training recommender systems. In this work, we aim to propose a probabilistic perspective for denoising implicit feedback, which has the ability to deal with several difficulties encountered by previous methods. 
In this section, we first describe the background of learning from binary or multi-class labels, from the perspective of implicit feedback recommendation and image classification, correspondingly. Then we provide a literate review about existing denoising methods.

\subsection{Learning from Binary Labels: Implicit Feedback Recommendation} % (fold)
\label{sub:implicit_feedback_usage_in_recommendation}
Implicit feedback recommendation is a typical learning field of binary labels. 
Modern recommender agents are usually trained using implicit feedback data, since the implicit data is much easier to be collected compared with explicit ratings. The binary implicit feedback data describes whether there was an interaction between the user and the item. A positive label denotes there were interactions while negative samples are usually sampled from missing interactions. Then both the positive samples and sampled negative examples are fed to perform pair-wise ranking \cite{rendle2014bayesian} or the binary cross-entropy (BCE) loss function. 
Besides, there are also attempts to investigate non-sampling approaches \cite{he2016fast,chen2019efficient,yuan2018fbgd} for implicit feedback. 
Regarding the recommendation models, MF \cite{koren2009matrix} is one of the most notable and effective models, which projects users and items to embedding vectors and then calculate the inner product between them as the prediction score. Recently, plenty of work has proposed deep learning-based recommendation models, such as GMF, NeuMF\cite{he2017neural}, CDAE\cite{wu2016collaborative} and Wide\&Deep \cite{cheng2016wide}. The key idea is to use deep learning to increase model expressiveness to capture more complex signals. Besides, graph neural networks also demonstrated their capability in recommendation.
Plenty of models have emerged, such as HOP-Rec \cite{yang2018hop}, KGAT \cite{wang2019kgat}, NGCF \cite{wang2019neural} and LightGCN \cite{he2020lightgcn}. This work aims at developing denoising methods which can be used to train various kinds of models. 

\subsection{Learning from Multi-class Labels: Image Classification} % (fold)
Image classification is one of the most important tasks in the field of computer vision which involves learning from multi-class labels. 
% \textcolor{red}{fill this section}.
Benchmarks like ImageNet\cite{deng2009imagenet} and CIFAR\cite{krizhevsky2009learning} have enormously promoted the development of image classification tasks, and have been well-studied to generate powerful AI systems, in which a large amount of work proposing various deep-learning based image classification models has emerged, such as InceptionNet\cite{szegedy2017inception}, ResNet\cite{he2016deep}, Vision-Transformer\cite{dosovitskiy2020image}, etc. However, as the expressive power of the models increases, the acquisition of large quantities of human-labeled images has become a frequent bottleneck in applying current image classification models. One simple solution is to turn to the crowdsourcing platforms, which however, could not guarantee the correctness of the collected labels. Thus in order to deploy the deep-learning models, effective denoising methods are of vital importance.

%. All of these methods, no matter model based or graph based, need implicit feedback as training set, thus they all will face the problem of noisy labels in implicit data. 
% subsection implicit_feedback_in_recommendation (end)
% \begin{figure}
% \centering     %%% not \center
% \subfigure[GMF on Modcloth]{\label{fig:GMF_modcloth_rating}\includegraphics[width=0.45\linewidth]{pictures/modcloth_ratings.png}}
% \subfigure[GMF on Electronics]{\label{fig:GMF_electronics_rating}\includegraphics[width=0.45\linewidth]{pictures/electronics_5_ratings.png}}
% % \subfigure[NeuMF on Electronics]{\label{fig:NeuMF_electronics_rating}\includegraphics[width=0.3\linewidth]{pictures/electronics_NeuMF_ratings.png}}
% \caption{Connection between Probabilities and Ratings. The first model name means the recommendation model, and the second model name indicates the probability model, i.e., $h$ in Eq.(\ref{h_model})}
% \label{Ratings}
% \end{figure}

\subsection{Literature Review of Denoising Methods } % (fold)
\label{sub:denoising_in_recommendation}
In recently years, there has been some researches focusing on denoising the corrupted labels. For example, in the recommendation field, some researches \cite{jagerman2019model,lu2018between} pointed out that the observed implicit feedback could be easily corrupted by different factors, such as popularity bias, conformity bias, exposure bias and position bias\cite{chen2020bias}. Training recommenders with corrupted implicit data would lead to misunderstanding of the real user preference and sub-optimal recommendation performance \cite{wang2020denoising, zhao2016gaze}. 
%Thus the interactions caused by the biases can't always reflect true user preferences and are considered as noisy examples As shown in \cite{wang2020denoising}, these noisy interactions (especially false positive ones) can drastically downgrade the performance of our recommender systems, while selecting negative samples randomly in training will also introduce noises and deteriorates the recommendation quality \cite{yuan2018fbgd}. 
Similar researches were also conducted on other machine learning application fields, often accompanied with the discussion of the generalization capability of deep learning models \cite{jiang2018mentornet}.

Existing denoising methods can be categorized into re-sampling methods \cite{yu2020sampler,gantner2012wbpr,ding2019samplerview,ding2018improvedview,ding2019reinforced,metalearning,autood}, re-weighting methods \cite{wang2020denoising,shen2019learning, nguyen2019self,debias} and methods utilizing additional knowledge input \cite{kim2014modeling,zhao2016gaze,lu2019effects}.
Re-sampling methods aim to design more effective samplers for data selection. For example, in the recommendation field, \cite{gantner2012wbpr} proposed to sample popular but not interacted items as negative examples while \citet{ding2018improvedview} proposed that the viewed but not purchased items are highly likely to be real negative. \citet{ding2019reinforced,wang2020reinforced} proposed to use reinforcement learning for negative sampling. 
\wy{Moreover, \cite{nguyen2019self, shen2019learning, song2020robust} also proposed well-known sampling strategies in more general machine learning field to filter out the noisy samples and try to resample the clean ones. }
The performance of re-sampling depends heavily on the sampling strategy \cite{yuan2018fbgd}, which is usually developed heuristically \cite{yu2020sampler}. 
Re-weighting methods 
usually identify the noisy examples as samples with high loss values and then assign lower weights to them. \wy{Representative works include \cite{shu2019meta} in the general machine learning field and \cite{wang2020denoising} in the recommendation field.}
But the denoising signal contained in loss values could encounter difficulties to distinguish noisy samples and hard clean samples. Curriculum learning and self-paced learning are also utilized to learn from corrupted labels. However, such kinds of methods usually need a clean dataset to train the teacher model or the meta-network \cite{jiang2018mentornet,shu2019meta}.
Additional knowledge such as dwell time\cite{kim2014modeling}, gaze pattern\cite{zhao2016gaze} and auxiliary item features\cite{lu2019effects} can also be used to denoise implicit feedback in the recommendation field. 
\xx{\cite{chen2021structured} proposed to improve the user-item graphs to avoid noisy connections. \cite{DenoisingUserAwareMemoryNetwork} proposed a feature purification module to denoise the dataset and then train a memory network upon that. The difference between their methods and ours is that they try to denoise the recommendation from the data perspective, while our methods focus on the model perspective, i.e., how to train a robust model from corrupted labels. \cite{IterativeRelabeledCF} proposes to iteratively relabel the dataset, which is more time-consuming. Besides, they only focus on noises in observed interactions, while our methods focus on noises in both click and not-click interactions when applying to the recommendation task. \cite{SGDL} proposes to use the losses in the early stages to find clean ones and then use them as the guiding signal for the following training. It's similar to the re-sampling methods and only considers the knowledge from a single model. }
However, such domain-specific knowledge is usually expensive to collect and cannot be generalized to other scenarios of corrupted data. 
\new{In addition, The methods mentioned above primarily consider the loss of samples as a measure of noisiness. For example, ITLM~\cite{shen2019learning} trims the dataset by selecting a certain percentage of the dataset with minimal loss. MW-Net~\cite{shu2019meta} incorporates the loss value into the network, using it to compute weights that are multiplied with the loss during training. T-CE~\cite{wang2020denoising} assigns zero weights to examples with large losses, employing a dynamic threshold. While the loss value is an important indicator for distinguishing between noisy and clean samples, it may not effectively differentiate between noisy samples and challenging yet clean samples. Notably, the loss value serves as the denoising signal in these methods, originating from a single model. In contrast, our approach derives the denoising signal from the differences observed across different models. It offers a new perspective of denoising and is robust to different kinds of noises and can be generalized to different settings and applications.}
\section{Motivation}
\label{sec:motivation}

In this section, we describe the empirical studies to illustrate the motivation of our denoising signal, which comes from the comparison between different model predictions on the same dataset. We describe the conducted studies on both binary implicit feedback data and multi-class image classification data.

\begin{figure*}
%\vspace{-12pt}
\centering
\subfigure[Different models.]{\label{fig:Differences_in_models}\includegraphics[width=0.40\linewidth]{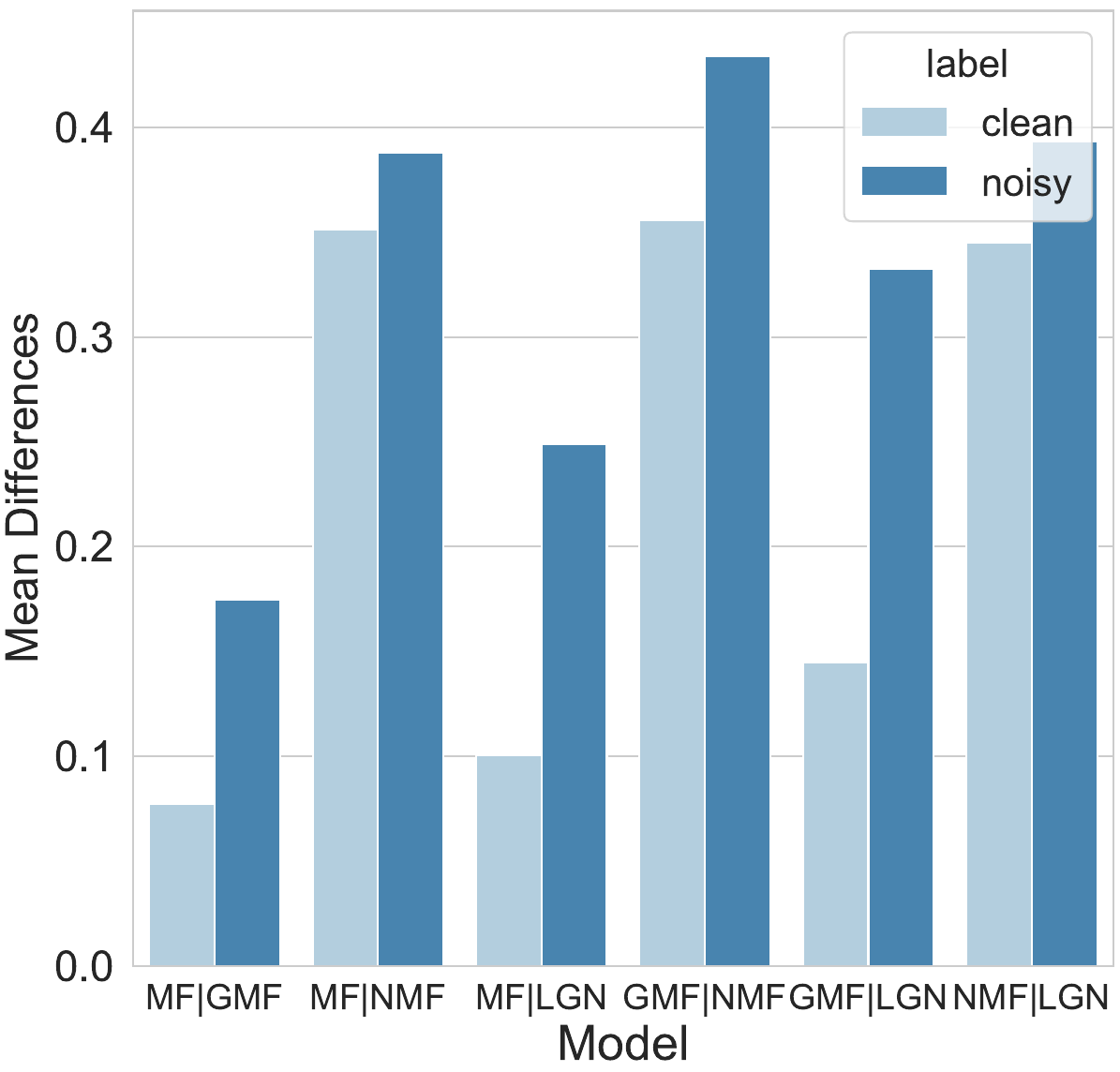}}
\subfigure[Different random seeds.]{\label{fig:Different_random_seeds}\includegraphics[width=0.40\linewidth]{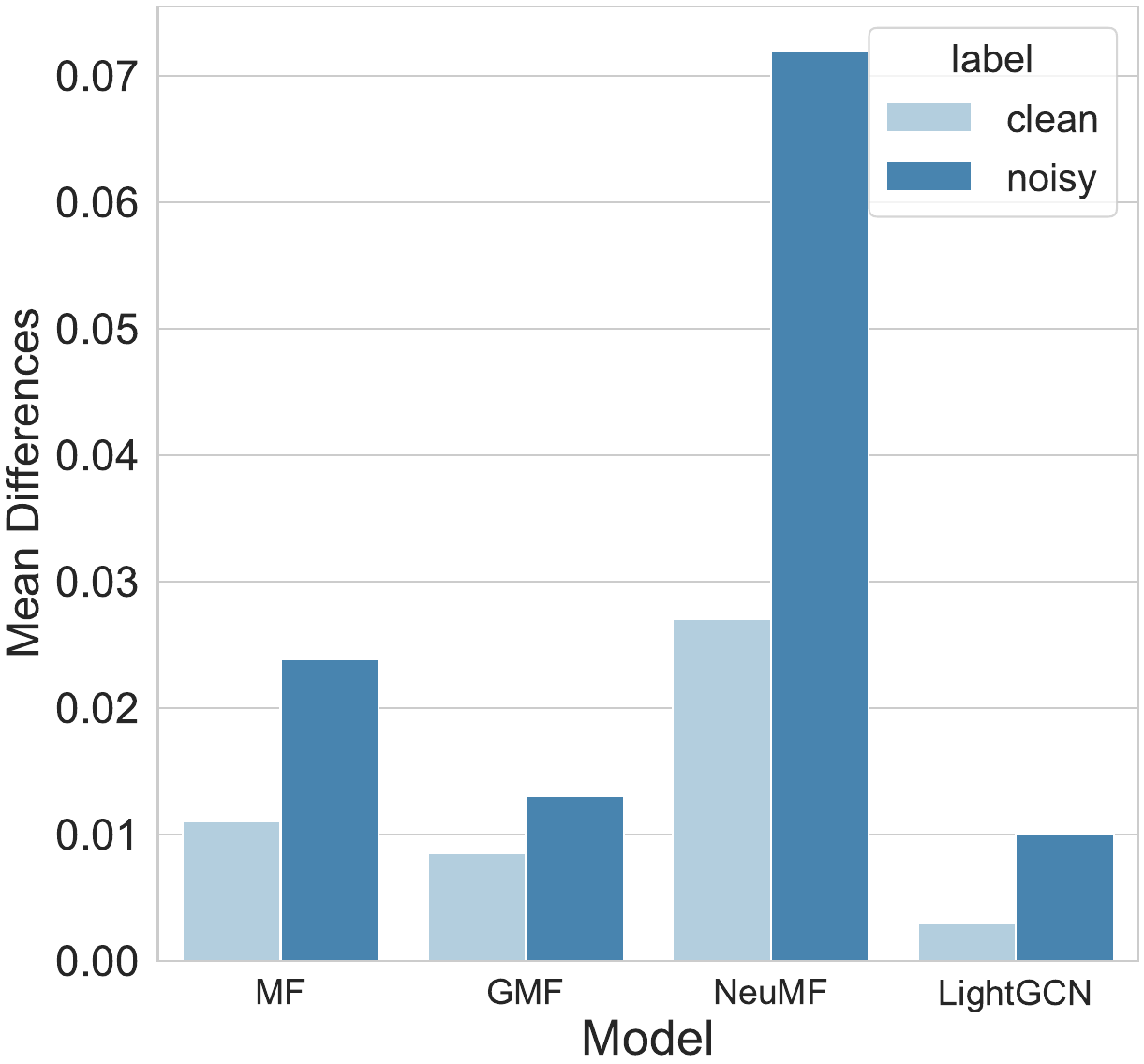}}
\vspace{-0.2cm}
\caption{Mean prediction differences on \textbf{clean} and \textbf{noisy} examples in ML-100k from (a) two different models or (b) one model trained with two different random seeds. NMF and LGN are short for NeuMF and LightGCN.}
\vspace{-0.2cm}
\label{fig:Analysis_on_memorizing_noisy_samples}
\end{figure*}

For binary implicit feedback data, we train four notable recommendation models (i.e. MF \cite{koren2009matrix}, GMF\cite{he2017neural}, NeuMF \cite{he2017neural} and LightGCN \cite{he2020lightgcn}) on the MovieLens 100K \footnote{https://grouplens.org/datasets/movielens/100k/}(ML-100K) dataset.
The user-item ratings are transferred to binary implicit feedback. All interacted items are considered as positive examples and negative examples are sampled from missing interactions.
Then among positive examples, we consider interactions whose ratings are 4 and 5 as clean (positive) examples while interactions whose ratings are 1 and 2 as noisy (positive) examples. \wy{Interactions with ratings equal to 3 are discarded to make the clean part and noisy part more separated and also make the results more convincing.
Same definitions can be found in \cite{wang2020denoising}.}
Figure \ref{fig:Differences_in_models} shows the mean 
prediction differences between two different models on clean examples and noisy examples. 
More precisely, the difference is defined as $|I(\delta(y_{ui}))-I(\delta(y'_{ui}))|$, where $y_{ui}$ and $y'_{ui}$ are the predicted scores from two models regarding user $u$ and item $i$, $\delta$ is the sigmoid function. $I(x)$ is an indicator function and is defined as $I(x)=1$ when $x\geq 0.5$ and $I(x)=0$ otherwise. 
We calculate all prediction differences on clean examples and noisy examples respectively and then report the average.
It's obvious that the model prediction differences %\zmc{The `prediction differences' is unclear here.} 
on noisy examples are significantly larger than the differences on clean examples. In other words, \emph{different models tend to make relatively similar predictions for clean examples compared with noisy examples}. Same results have also been found in several other binary datasets.  That is to say, different models tend to fit different parts of the corrupted data but clean examples are the robust component which every model attempts to fit. This observation also conforms with the nature of robust learning. %Nevertheless, for noisy examples, the predictions from different models vary much more and are of higher variance. 
Besides, we also find that even one model trained on the same dataset with different random seeds tends to make more consistent predictions on clean examples compared with noisy examples, as shown in Figure \ref{fig:Different_random_seeds}. 

\begin{figure*}
%\vspace{-15pt}
\centering
\subfigure[Different models.]{\label{fig:Differences_in_models_m}\includegraphics[width=0.40\linewidth]{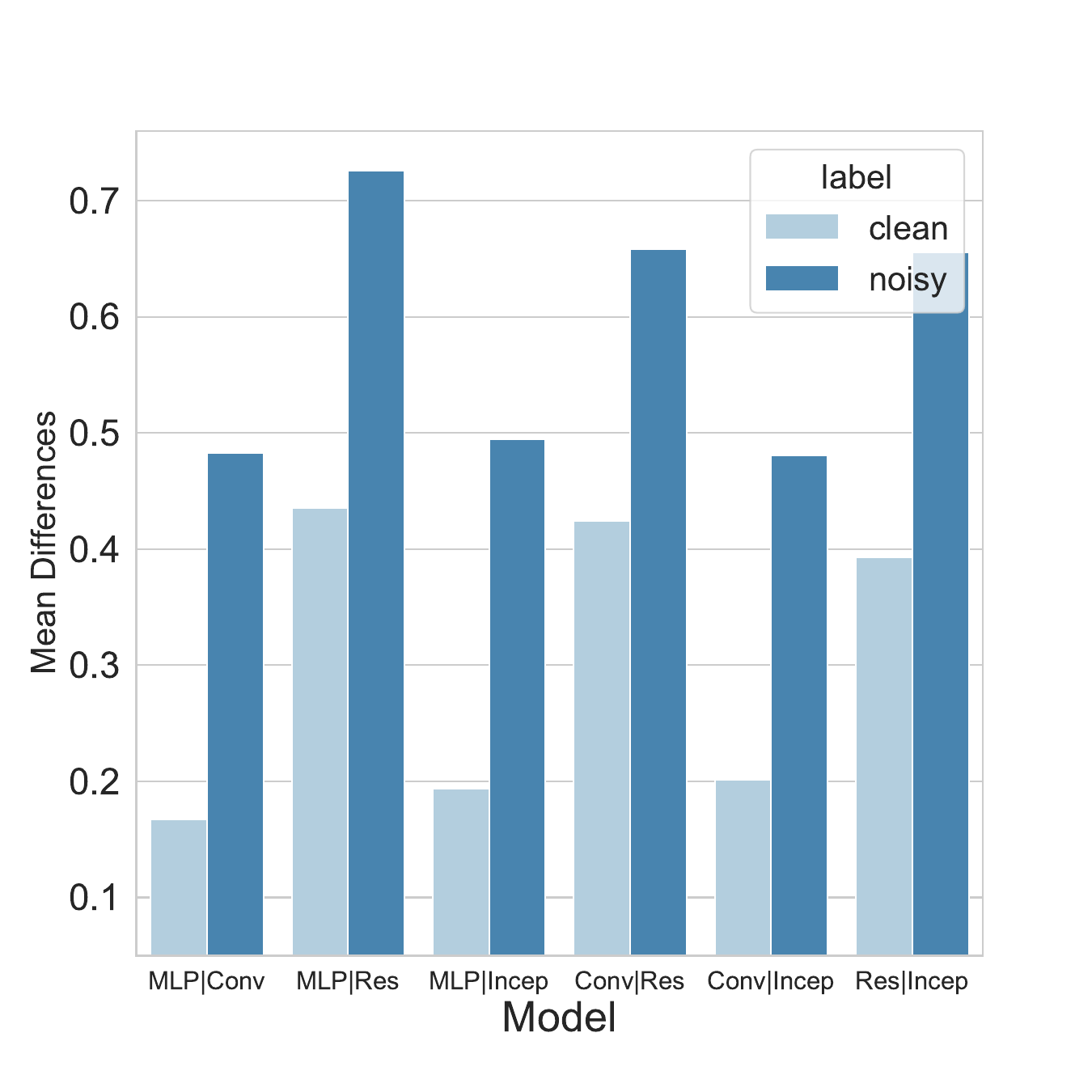}}
\subfigure[Different random seeds.]{\label{fig:Different_random_seeds_m}\includegraphics[width=0.40\linewidth]{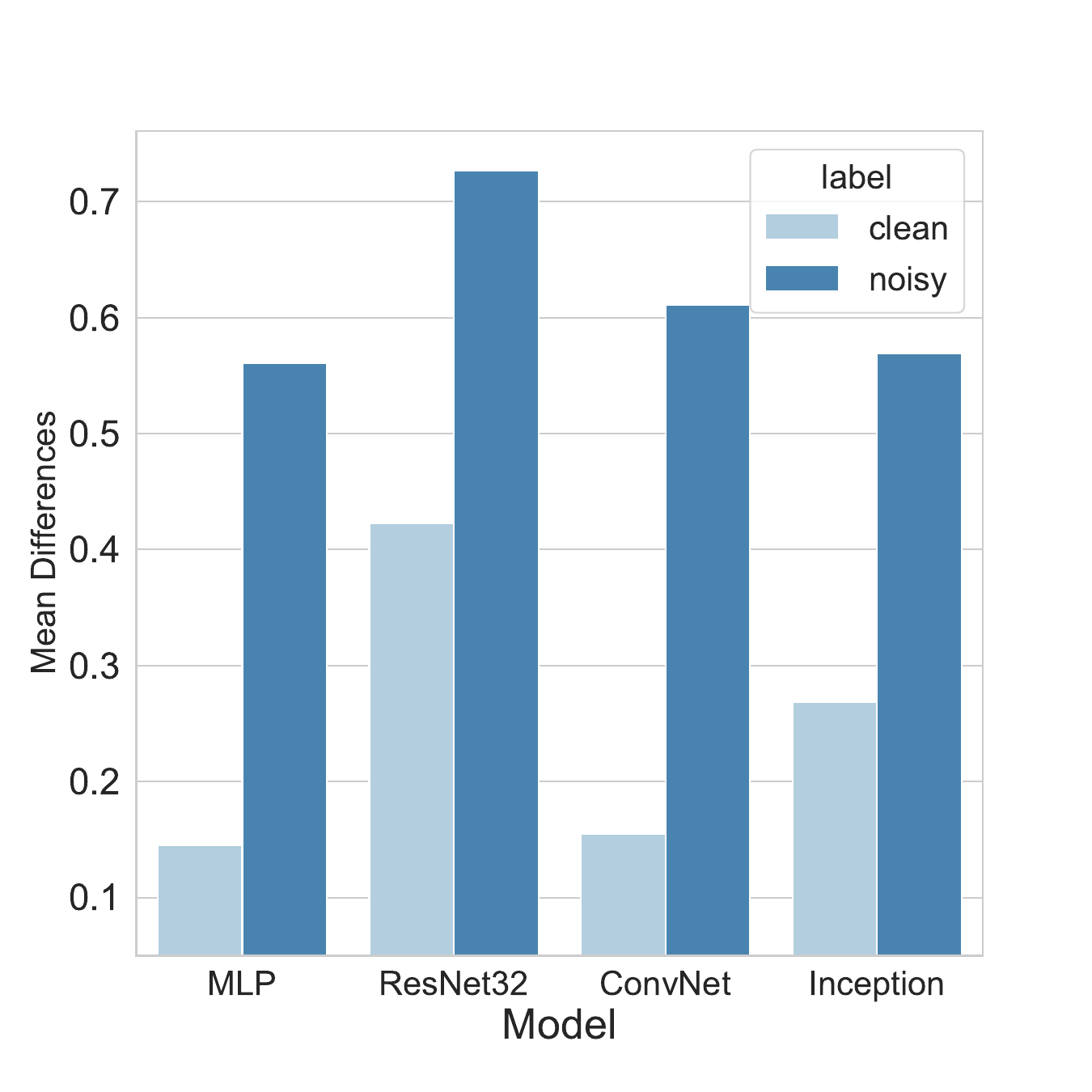}}
\vspace{-0.2cm}
\caption{Mean prediction differences on \textbf{clean} and \textbf{noisy} examples in MNIST from (a) two different models or (b) one model trained with two different random seeds. Conv, Res and Incep is short for ConvNet and ResNet and InceptionNet respectively.}
% \vspace{-0.2cm}
\label{fig:Analysis_on_memorizing_noisy_samples_mnist}
\end{figure*}
For multi-class data, we choose the benchmark MNIST\footnote{\url{https://pytorch.org/vision/stable/datasets.html\#mnist}} dataset. We corrupted the original data by replacing 40\% of correct labels with random labels. The corrupted samples are considered as noisy instances while the original ones are treated as clean samples. The same operation was also adopted in \cite{shu2019meta,shen2019learning}.
We then train two well-known image classification models: ResNet32\cite{he2016deep}, and InceptionNet\cite{szegedy2017inception}. We also train a simple 4-layer multi-layer perceptron (MLP) and a 2-layer Convolution network as a baseline. The prediction difference between two models on one instance $\mathbf{X}_i$ is calculated as $I'(M_1(\mathbf{X}_i)==M_2(\mathbf{X}_i))$, where $M_1$ and $M_2$ denote the classification results of the two models, respectively. $I'(x)$ is an indicator function with $I'(x)=1$ if $x$ is true and $I'(x)=0$ otherwise.
Then we calculate the prediction differences on all instances of the test set and report the mean difference on clean samples and noisy samples separately.
We can see from Figure \ref{fig:Analysis_on_memorizing_noisy_samples_mnist} that same observations also exit in multi-class data. Different models tend to make similar predictions on clean instances while predictions on noisy samples vary much more across different models. This common observation in both binary data and multi-class data serves as our motivation to design the denoising method. 

\section{METHODOLOGY} % (fold)
\label{sec:method}
In this section, we propose the learning framework DeCA which uses the observations described in section \ref{sec:motivation} as the denoising signals to learn robust machine learning models. 
% We first describe the denoising for binary implicit feedback recommendation, then we generalize the method for multi-class image classification.
\wy{We start with the binary classification problem, and then extend it into a multi-class classification scenario.}

\wy{
\subsection{Denoising for Binary Classification}
Before the detailed description of the methods,  we first present some notations and problem formulation.
\subsubsection{Notations and Problem Formulation}
\label{sub:notations_and_problem_formulation}
The dataset is denoted as $\{x_i, \tilde{y}_i\}_{i=1}^N$, where $N$ is the total number of samples and $\tilde{y}_i$ represents the corrupted binary labels. The ground truth labels are denoted as $\{y_i\}_{i=1}^N$, which are inaccessible during training. The samples and corrupted labels in the dataset are represented as $\textbf{X} \in \mathbb{R}^{N*K}$ and $\tilde{\textbf{Y}} \in \mathbb{R}^{N}$, respectively, where $K$ is the feature dimension. Similarly, the ground truth labels are denoted as $\textbf{Y} \in \mathbb{R}^{N}$. It should be noted that $\textbf{Y}$ and $\tilde{\textbf{Y}}$ are not identical due to the presence of noise in the data. We make the following assumption:
\begin{assumption} \label{bernoulli_assumption}
We assume $y_i$ is drawn from a Bernoulli distribution:
\begin{equation}\label{y-berno}
    y_i \sim Bernoulli(\eta_i)\approx Bernoulli(f_\theta(x_i)),
\end{equation} 
\end{assumption}
where $\eta_i$ describes the probability of positive class (i.e., $y_i=1$) and is approximated by $f_\theta(x_i)$. $\theta$ denotes the parameters of $f$. We use $f_\theta$ as our target model that generates the predictions. Specifically, the output value after sigmoid of the logits from the model is denoted as $f_\theta(x_i)$. \wyy{Please be advised that the probability denoted by $P(y_i)$ above, as well as all subsequent probabilities discussed in this paper, are conditioned on X. Therefore, in the interest of simplicity, \textbf{we shall exclude X from the expressions utilized in our deductions} involving these distributions.}
%which is also the probability that $\eta_i$ equals to $1$. 
Generally speaking, $f_\theta$ is expected to have high model expressiveness. Since the denoising singal comes from the agreement across different model predictions, we introduce another auxiliary Bernoulli distribution parameterized by $g_\mu(x_i)$, which is another model and also built for the classification task. 
Considering both the noisy positive and noisy negative examples, we then give the following assumption: 

\begin{assumption}\label{h_model_assumption_for_binary}
  we assume that given the underlying true label $y_i$, the corrupted binary label $\tilde{y}$ is also drawn from Bernoulli distributions as: 
\begin{equation}\label{h_model}
  \begin{array}{c}
  \tilde{y}_{i}|y_{i}=0, x_i \sim Bernoulli(h_\phi(x_i))\\
  \tilde{y}_{i}|y_{i}=1, x_i \sim Bernoulli(h'_\psi(x_i)),
  \end{array}
\end{equation}
where $h_\phi(x_i)$ and $h'_\psi(x_i)$, parameterized by $\phi$ and $\psi$ respectively, are two models describing the consistency between the underlying true labels and the noisy labels in the dataset.   
\end{assumption}

Our task is, given the corrupted binary labels $\tilde{\textbf{Y}}$ and the input dataset $\textbf{X}$, we want to infer the true labels $\textbf{Y}$ and the underlying model $f_\theta$. Then $f_\theta$ is used for model inference and predictions.

\subsubsection{Denoising with Cross-Model Agreement}
\label{ssub:denoising_with_cross_model_agreement}
As discussed in Section \ref{sec:motivation}, for clean examples which denote the true label distribution, different models tend to make more consistent predictions compared with noisy examples. For simplicity, in this subsection we use $P(\mathbf{Y})$ to denote the true label distribution $Bernoulli(\eta)$. $P_f(\textbf{Y})$ and $P_g(\textbf{Y})$ are used to represent the approximated $Bernoulli(f_\theta)$ and $Bernoulli(g_\mu)$, correspondingly.
Due to the fact that $P_f(\textbf{Y})$ and $P_g(\textbf{Y})$ both approximate the true label  $\textbf{Y}$, they should remain a relatively small KL-divergence according to section \ref{sec:motivation}, which is formulated as 
\begin{equation}
\label{eq:kl-original}
D[P_g(\textbf{Y})||P_f(\textbf{Y})]=E_{\textbf{Y}\sim P_g}[\log P_g(\textbf{Y})-\log P_f(\textbf{Y})].
\end{equation}
However, naively optimizing Eq.(\ref{eq:kl-original}) is meaningless since we do not have the supervision signal of $\mathbf{Y}$. As a result, we need to introduce supervision signals from the corrupted data observation $\tilde{\textbf{Y}}$. Using the Bayes theorem, $P_f(\textbf{Y})$ can be approximated as 
\begin{equation}
\label{eq:bayesian}
P_f(\textbf{Y})\approx P(\textbf{Y})=\frac{P(\tilde{\textbf{Y}}) P(\textbf{Y}|\tilde{\textbf{Y}})}{P(\tilde{\textbf{Y}}|\textbf{Y})}.
\end{equation}
Combining Eq.(\ref{eq:bayesian}) and Eq.(\ref{eq:kl-original}), we can obtain
\begin{align}
\label{eq:transformation}
   D&[P_g(\textbf{Y}) || P_f(\textbf{Y})] = E_{\textbf{Y} \sim P_g}[\log P_g(\textbf{Y}) - \log P_f(\textbf{Y})] \nonumber \\
  &\approx E_{\textbf{Y} \sim P_g} [\log P_g(\textbf{Y}) - \log \frac{P(\tilde{\textbf{Y}}) P(\textbf{Y}|\tilde{\textbf{Y}})}{P(\tilde{\textbf{Y}}|\textbf{Y})}] \nonumber \\
  &= E_{\textbf{Y} \sim P_g}[\log P_g(\textbf{Y}) - \log P(\textbf{Y}|\tilde{\textbf{Y}}) - \log P(\tilde{\textbf{Y}}) + \log P(\tilde{\textbf{Y}}|\textbf{Y})]  \nonumber \\
  &= D[P_g(\textbf{Y})||P(\textbf{Y}|\tilde{\textbf{Y}})] - \log P(\tilde{\textbf{Y}}) + E_{\textbf{Y} \sim P_g}[\log P(\tilde{\textbf{Y}}|\textbf{Y})].
\end{align}
We then rearrange the terms in Eq.(\ref{eq:transformation}) and obtain
\begin{align}
  E_{\textbf{Y} \sim P_g}[\log P(\tilde{\textbf{Y}}|\textbf{Y})] &- D[P_g(\textbf{Y}) || P_f(\textbf{Y})] \nonumber \\
  &= \log P(\tilde{\textbf{Y}}) - D[P_g(\textbf{Y})||P(\textbf{Y}|\tilde{\textbf{Y}})]. \label{lower_bound}
\end{align}
We can see the meaning of maximizing the left side of Eq.(\ref{lower_bound}) is maximizing the likelihood of data observation given underlying true labels (i.e., $\log P(\tilde{\textbf{Y}}|\textbf{Y})$) and meanwhile minimizing the KL-divergence between two models which both approximate the true labels (i.e., $D[P_g(\textbf{Y}) || P_f(\textbf{Y})]$).
Since the KL-divergence $D[P_g(\textbf{Y})||P(\textbf{Y}|\tilde{\textbf{Y}})]$ is larger than zero, the left side of Eq.(\ref{lower_bound}) can also be seen as the lower bound of $\log P(\tilde{\textbf{Y}})$. The bound is satisfied only if $P_g(\textbf{Y})$ perfectly recovers $P(\textbf{Y}|\tilde{\textbf{Y}})$, in other words, $P_g(\textbf{Y})$ perfectly approximates the underlying true labels given the corrupted data.

A naive solution is to directly maximize the left side of Eq.(\ref{lower_bound}) with an end-to-end fashion. However, it would not yield satisfactory performance. The reason is that the left side of Eq.(\ref{lower_bound}) is based on the expectation over $P_g$. The learning process is equivalent to training $g_\mu$ with the corrupted data $\tilde{\textbf{Y}}$ and then uses $D[P_g(\textbf{Y})||P_f(\textbf{Y})]$ to transmit the information from $g_\mu$ to $f_\theta$. However, such the learning process is ineffective to train the target model $f_\theta$. To fix the problem, we notice that when the training process is converged, two distributions $P_f(\textbf{Y})$ and $P_g(\textbf{Y})$ would be close to each other. So we can then modify the left side of Eq.(\ref{lower_bound}) as
\begin{align}
  \label{trick}
  E_{\textbf{Y} \sim P_f}[\log P(\tilde{\textbf{Y}}|&\textbf{Y})] - D[P_g(\textbf{Y}) || P_f(\textbf{Y})] \nonumber \\
  &\approx \log P(\tilde{\textbf{Y}}) - D[P_g(\textbf{Y})||P(\textbf{Y}|\tilde{\textbf{Y}})].
\end{align}
Then optimizing the left side of Eq.(\ref{trick}) is actually training $f_\theta$ with the corrupted $\tilde{\textbf{Y}}$ and then transmit information to $g_\mu$. However,  $g_\mu$ could only fit the robust data component (i.e., clean examples). 
Thus $g_\mu$ would not affect the learning of $f_\theta$ on clean examples, while in the meantime, pull back $f_\theta$ on noisy samples, or in other words, downgrade the noisy signal. To this end, the denoising objective function can be formulated as
\begin{align}
  \label{L_simple}
  \mathcal{L}=-E_{\textbf{Y} \sim P_f}[\log P(\tilde{\textbf{Y}}|\textbf{Y})]+D[P_g(\textbf{Y}) || P_f(\textbf{Y})].
\end{align}
Considering that the gradient of $D[P_g||P_f]$ and $D[P_f||P_g]$ to $\theta$ is different, which could slightly affect the performance, we then formulate the final denoising objective function as
\begin{equation}
  \mathcal{L}_{DeCA} = -E_{\textbf{Y}\sim P_f}[\log P(\tilde{\textbf{Y}}|\textbf{Y})] + \alpha D[P_g||P_f] + (1-\alpha)D[P_f||P_g],
  \label{DPI-loss}
\end{equation}
where $\alpha \in [0,1]$ is a hyper-parameter.
In fact, the two models $f_\theta$ and $g_\mu$ in Eq.(\ref{DPI-loss}) can play a symmetric role, but we still need a model (here we use $f_\theta$) to do the inference. Thus, the other model (here we use $g_\mu$) serves as an auxiliary model. 

The detailed formulation of the term $E_{\textbf{Y}\sim P_f}[\log P(\tilde{\textbf{Y}}|\textbf{Y})]$ in $\mathcal{L}_{DeCA}$ is derived as: 

\begin{align}
\small
  &E_{\textbf{Y}\sim P_f}[\log P(\tilde{\textbf{Y}}|\textbf{Y})] = \sum_{i} E_{r_{i}\sim P_f} [\log P(\tilde{y}_{i}|y_{i})] \nonumber \\
  &=\sum_{i|\tilde{y}_i=1}  \left\{\begin{array}{ll}
    \log P(\tilde{y}_i=1|y_i=1) \cdot P_f(y_i=1) \\+
     \log P(\tilde{y}_i=1|y_i=0)\cdot P_f(y_i=0) \end{array}\right. \nonumber\\
     &+\sum_{(u,i)|\tilde{y}_i=0} \left\{\begin{array}{ll}
    \log P(\tilde{y}_i=0|y_i=1) \cdot P_f(y_i=1) \\
    + \log P(\tilde{y}_i=0|y_i=0)\cdot P_f(y_i=0)
  \end{array}\right. \nonumber \\
  &=\hspace{-0.3cm}\sum_{i|\tilde{y}_i=1}
    \log h'_\psi(i) \cdot f_{\theta}(x_i) + \log h_\phi(i) \cdot (1-f_\theta(x_i))\nonumber \\
  &+\hspace{-0.3cm}\sum_{i|\tilde{y}_i=0}
  \log(1 - h'_\psi(x_i))\cdot f_{\theta}(x_i) + \log(1 - h_\phi(x_i) ) \cdot (1-f_\theta(x_i)).
  \label{eq:likelihood}
\end{align}
The KL divergence term $D[P_g||P_f]$ can be calculated as
\begin{equation*}
  D[P_g||P_f] = g_\mu(x_i) \cdot \log \frac{g_\mu(x_i)}{f_\theta(x_i)} + (1-g_\mu(x_i)) \cdot \log \frac{1 - g_\mu(x_i)}{1-f_\theta(x_i)}.
\end{equation*}
$D[P_f||P_g]$ is computed similarly with $D[P_g||P_f]$.

\subsubsection{DeCA with Fixed Pre-training}
\label{ssub:deca_with_fixed_pre_training}
As described in subsection \ref{ssub:denoising_with_cross_model_agreement}, DeCA utilizes an auxiliary model $g_\mu$ as an information filter which helps to downgrade the effect of noisy signal. The auxiliary model $g_\mu$ is co-trained jointly with the target model $f_\theta$. As discussed before, the left side of Eq.(\ref{lower_bound}) is the lower bound of $\log P(\tilde{\textbf{Y}})$, which is satisfied when $P_g(\mathbf{Y}) \approx P(\textbf{Y}|\tilde{\textbf{Y}})$. Then as the training process converge, we have $P_g(\mathbf{Y}) \approx P_f(\mathbf{Y}) $. Thus we could expect $P(\textbf{Y}|\tilde{\textbf{Y}}) \approx P_f(\mathbf{Y})$.
To this end, we modify the assumption of Eq.(\ref{y-berno}) as 
\begin{equation}
\label{rui|r-berno}
  y_{i}|\tilde{y}_{i} \sim Bernoulli(\eta_{i})\approx Bernoulli(f_\theta(x_i)).
\end{equation} 
The underlying intuition is that whether the data observation is known or not should not affect the robust underlying true labels, which is reasonable and keeps inline with the nature of robust learning. 
As a result, we should minimize the following KL-divergence
\begin{equation}\label{eq:kl}
    D[P_f(\textbf{Y}|\tilde{\textbf{Y}}) || P(\textbf{Y}|\tilde{\textbf{Y}})].
\end{equation}
Through Bayesian transformation, the following equation can be deduced:
\begin{equation}
\label{VAE_lower_bound}
E_{P_f}[\log P(\tilde{\textbf{Y}}|\textbf{Y})] - D[P_f(\textbf{Y}|\tilde{\textbf{Y}})|| P(\textbf{Y})] \nonumber = \log P(\tilde{\textbf{Y}}) - D[P_f(\textbf{Y}|\tilde{\textbf{Y}})||P(\textbf{Y}|\tilde{\textbf{Y}})].
\end{equation}
Similarly with Eq.(\ref{DPI-loss}), we propose another variant loss function as DeCA with pre-training (DeCA(p)): 
\begin{align}
\label{DVAE-loss}
  \mathcal{L}_{DeCA(p)} &= - E_{P_f}[\log P(\tilde{\textbf{Y}}|\textbf{Y})] + \alpha D[P_f(\textbf{Y}|\tilde{\textbf{Y}}) || P(\textbf{Y})] + (1- \alpha)D[P(\textbf{Y}) || P_f(\textbf{Y}|\tilde{\textbf{Y}})].
\end{align}

We use a pre-trained model $f_{\theta'}$ which has the same structure as our target model $f_\theta$ but is trained with different random seeds to model $P(\mathbf{Y})$. This setting is motivated by the observation that one model trained with different random seeds tends to make high variance predictions on noisy examples but more consistent agreement predictions on clean examples, as shown in Figure \ref{fig:Different_random_seeds}. 

The major difference between DeCA and DeCA(p) can be summarized as follows:
\begin{itemize}[leftmargin=*]
  \item The auxiliary model $g_\mu$ of DeCA could be any model  while the involved $f_{\theta'}$ in DeCA(p) has the same model structure as the target model $f_\theta$.
  \item DeCA co-trains the auxiliary model $g_\mu$ with $f_\theta$ but DeCA(p) uses a pre-trained and fixed $f_{\theta'}$ to describe the prior distribution.
\end{itemize}

\subsection{Denoising for Multi-Class Classification}
% \section{Method}
In this section, we extend the proposed DeCA methods from the binary scenario to the multi-class scenario.

\subsubsection{Notations and Problem Formulation}
Similarly to Section \ref{sub:notations_and_problem_formulation}, the dataset is still denoted as $\{x_i, \tilde{y}_i\}_{i=1}^N$, with $\{y_i\}_{i=1}^N$ being the ground truth labels. $\textbf{X}, \textbf{Y}, \tilde{\textbf{Y}}$ are the features, underlying true labels, observed corrupted labels, respectively. Here we \wyy{denote all the possible classes as $\mathcal{C} \{0,\cdots,|\mathcal{C}|-1\}$} and now we have $\tilde{y}_i, y_i \in \mathcal{C}$. 
Assumption \ref{bernoulli_assumption} is then updated to:
\begin{assumption}\label{multinomial_assumption}
True label $y_i$ is drawn from a multinomial distribution conditioned on $x_i$ and is not dependent on $\tilde{y}_i$:
\begin{equation}\label{encoder}
    y_i \sim Mult(|\mathcal{C}|, \eta_i) \approx Mult(|\mathcal{C}|, f_\theta(x_i))
\end{equation}
where $\eta_i \in \mathbb{R}^{C}$ describes the probability distribution of $i$-th example over all $|\mathcal{C}|$ classes and is approximated by $f_\theta(x_i)$. $\theta$ denote the parameter of function $f$. $Multi(|\mathcal{C}|, \eta_i)$ refers to the multinomial distribution with probabilities $\eta_i$. 
\end{assumption}
Besides, Assumption \ref{h_model_assumption_for_binary} is updated to: 
\begin{assumption}
Given true label $y_i$ and the feature $x_i$, the observed noisy label $\tilde{y}_i$ is also drawn from a multinomial distribution:
\begin{equation}\label{decoder}
    \tilde{y}_i | x_i, y_i \sim Mult(|\mathcal{C}|, h_\varphi(x_i, y_i)) 
\end{equation}
where $h_\varphi$ is the model representing the mapping from $(x_i, y_i)$ to $\tilde{y}_i$, and $\phi$ is the parameters of the model. With $h_\varphi$ we are able to describe the correlation between $x_i, y_i$ and $\tilde{y}_i$. Note that we combine two models $h_\phi$ and $h'_\psi$ in Eq.(\ref{h_model}) into a single model $h_\varphi$ but conditioned on $y_i$. 
\end{assumption}

\subsubsection{DeCA for the Multiple Class Scenario.}
Similarly with Section \ref{ssub:deca_with_fixed_pre_training}, we also need to minimize the following KL-divergence:
\begin{equation*}
    D[P_f(\textbf{Y}|\tilde{\textbf{Y}}) || P(\textbf{Y}|\tilde{\textbf{Y}})].
\end{equation*}
Then with Bayesian transformation (the same logic with Section \ref{ssub:deca_with_fixed_pre_training}), we have:
\begin{equation*}
    E_{P_f}[\log P(\tilde{\textbf{Y}}|\textbf{Y})] - D[P_f(\textbf{Y}|\tilde{\textbf{Y}})|| P(\textbf{Y})] \nonumber = \log P(\tilde{\textbf{Y}}) - D[P_f(\textbf{Y}|\tilde{\textbf{Y}})||P(\textbf{Y}|\tilde{\textbf{Y}})].
\end{equation*}
Then we extract the left side of the last equation as our objective here:
\begin{equation}\label{multi-class-loss}
      \mathcal{L}_{DeCA(p)} = - E_{P_f}[\log P(\tilde{\textbf{Y}}|\textbf{Y})] + D[P_f(\textbf{Y}|\tilde{\textbf{Y}}) || P(\textbf{Y})]
\end{equation}
Here $P(\textbf{Y})$ is the prior distribution, which will be modeled by a pretrained model $f_{\theta'}$. $f_{\theta'}$ has the same structure as our target model $f_\theta$ but is trained with a different random seed, similar to the setting in Section \ref{ssub:deca_with_fixed_pre_training}. 
Compared to the binary classification setting, there are two major differences: \\
(1) The detailed formulation of the term $E_{P_f}[\log P(\tilde{\textbf{Y}}|\textbf{Y})]$ in $\mathcal{L}$ is:
\begin{equation}\label{multiclass-first-term}
    E_{{\textbf{Y}\sim P_f}}[\log P(\tilde{\textbf{Y}} | \textbf{Y})] = \sum_i \sum_{c\in \mathcal{C}} \log P(\tilde{y}_i|y_i = c) * P(y_i = c)
\end{equation}
where $P(\tilde{\textbf{Y}} | \textbf{Y})$ is given by $h_\varphi$ defined in Eq.(\ref{encoder}). \\
(2) The KL divergence term $D[P_f(\textbf{Y}|\tilde{\textbf{Y}}) || P(\textbf{Y)}]$ is calculated as:
\begin{equation}
    D[P_f(\textbf{Y}|\tilde{\textbf{Y}}) || P(\textbf{Y})] = \sum_{c\in \mathcal{C}} f_\theta(x_i, c) \cdot \log \frac{f_\theta(x_i, c)}{f_{\theta'}(x_i, c) }
\end{equation}
where $f_\theta(x_i, c)$ is the probability predicted by $f_\theta$ of $x_i$ being in class $c$. 
%With this two differences, we could extend DeCA(p) into multi-class scenario. 

\subsection{Training Routine}
\label{ssec:training_routine}
We can see there is a expectation term in Eq.(\ref{DPI-loss}), Eq.(\ref{DVAE-loss}), Eq.(\ref{multi-class-loss}), which is $E_{P_f} [\log P(\tilde{\textbf{Y}}|\textbf{Y})]$. A naive solution to optimize $E_{P_f} [\log P(\tilde{\textbf{Y}}|\textbf{Y})]$ could be performing the calculations in the objectives and back-propagate directly. However, In our implementation, it won't yield satisfactory results. 
The reason is that computing $E_{P_f} [\log P(\tilde{\textbf{Y}}|\textbf{Y})]$ requires computing $P(\tilde{\textbf{Y}} | \textbf{Y})$ (or $h$-models ($h_\phi$ and $h_\psi'$) in Eq.(\ref{h_model}) and $h_\varphi$ in Eq.(\ref{encoder}) firstly. 
However, in the early stage of training, the true labels $\textbf{Y}$ is hardly inferred by the under-trained model $f_\theta$, and the $h$-models themselves are also corrupted. Then the objective  Eq.(\ref{DPI-loss}), Eq.(\ref{DVAE-loss}), Eq.(\ref{multi-class-loss}) will fail to refine the model $f_\theta$ and the $h$-models. Instead, they would stuck in poor local minimum where the loss is small but $f_\theta$ and $h$-models are meaningless.
To handle this issue, we design an iterative training routine to perform more effective learning. We elaborate the routine for multi-class scenario, and setting the number of classes to two can easily lead to the formulation in binary-class settings. 
For the model $h_\varphi$ in Eq.(\ref{decoder}), the input $y_i$ has $|\mathcal{C}|$ possible values. We force the model to attend to one value in each timestep.

\noindent At timestep $T$, let $k = T$ mod $|\mathcal{C}|$. In this step, $\forall c \in \mathcal{C}$ we set the probability $P(\tilde{y}_i|y_i)$ as: 

\begin{equation}\label{training_routine_set_conditional_probability}
    P(\tilde{y}_i = c|y_i = c) = \left\{\begin{array}{ll}
        1, & c\neq k \\
        h_\varphi(x_i, y_i, c), & c=k \\
    \end{array}\right.
\end{equation}
where $h_\varphi(x_i, y_i, c) = $ the $c$-th probability of $h_\varphi(x_i, y_i)$ for $c\in\mathcal{C}$. 

Once Eq.(\ref{training_routine_set_conditional_probability}) is given, the following several equations are yielded:
\begin{align*}
    & P(\tilde{y}_i = k|y_i\neq k) = 0; \\
    & P(\tilde{y}_i \neq k, \tilde{y}_i \neq y_i |y_i\neq k) = 0; \\
    & P(\tilde{y}_i \neq k|y_i =k) = h_\varphi(x_i, y_i, \tilde{y}_i)
\end{align*}
\wyy{In the following part, we denote $C_{k1}$ and $C_{k2}$ as follows:
\begin{align}
    C_{k1} = -\log P(\tilde{y}_i = k|y_i\neq k) \label{eq:C_k1} \\
    C_{k2} = - \log P(\tilde{y}_i \neq k, \tilde{y}_i \neq y_i |y_i\neq k) \label{eq:C_k2}
\end{align}}
The reason of doing this is that these two variables will appear in the following deductions and $-\log 0$ are generally not calculable. So we need to find some large positive numbers to substitute them. There are several points here to note: (1) $C_{k1}$ and $C_{k2}$ shrink to only one parameter $C_k$ when in binary class classification problems because it's not possible to have the case that $\tilde{y}_i \neq k, \tilde{y}_i \neq y_i, y_i\neq k$. So there are two parameters in binary classification problems: $C_k, k=1,2$. (2) We define $C_1$ and $C_2$ here because we want to pay extra attention on class $k$ at the current step, so when $\tilde{y}_i \neq k$, we would like to use a specific pamameter $C_{k1}$ to denote $-\log P(\tilde{y}_i = k| y_i \neq k)$. So for each class, we have two parameters $C_{k1}$ and $C_{k2}$ here. So our framework could potentially be applied to various kinds of noises.  

\noindent With the above equations, Eq.(\ref{multiclass-first-term}) becomes:

\begin{align}
E_{\textbf{Y}\sim P_f}[\log P(\tilde{\textbf{Y}}|\textbf{Y})] 
&= E_{Y\sim P_f} \sum_i P(y_i=c) \log P(\tilde{y}_i|y_i=c) \nonumber \\
&= \sum_{i|\tilde{y}_i=k} \sum_{c\in \mathcal{C} \& c\neq k} -C_{1k} * P(y_i=c) + \log P(\tilde{y}_i=k|y_i=k) * P(y_i=k) \nonumber \\
& + \sum_{i|\tilde{y}_i\neq k}  \sum_{c \in \mathcal{C} \& c \neq k, c\neq \tilde{y}} -C_{2k}*P(y_i=c) + \log(P(\tilde{y}_i|y_i=k)) P(y_i = k)  \nonumber \\
& = \sum_{i|\tilde{y}_i=k} - C_{1k} * (1-P(y_i = k)) + \log h_\varphi(x_i, k, k) * P(y_i=k) \nonumber \\
& + \sum_{i|\tilde{y}_i\neq k}  -C_{2k} * (1 - P(y_i = k) - P(y_i = \tilde{y}_i)) + \log(h_\varphi(x_i, k, \tilde{y}_i)) P(y_i = k)  \label{training_objective_1}
\end{align}
where $C_{1k}$ is the large positive hyper-parameter used to substitute the $-\log (P(\tilde{y}_i = k | y_i = c)), \forall c \in \mathcal{C} \backslash \{k\}$, and $C_{2k}$ is another large positive hyper-parameter representing $-\log (P(\tilde{y}_i| y_i = c)), \forall c \in \mathcal{C} \backslash \{k, \tilde{y}_i\}$. Intuitively, we constrain all the correlations between $c\neq k$, $y$ and $\tilde{y}$ but leave the function lying between $\tilde{y}$ and $k$ unconstrained, denoted by $P(\tilde{y}_i|y_i=k)$. Thus the objective of current step is to learn the part that is only correlated with $y_i = k$ in $h_\varphi$. 

\noindent After certain epochs (the number of epochs is a hyperparameter and is usually chosen at the point that the above part is converged), when the probability distributions parameterized by $h_\varphi$ have converged, we can remove $C_{k1}$ and $C_{k2}$ and change Eq.(\ref{training_routine_set_conditional_probability}) to:
\begin{equation}\label{training_routine_set_conditional_probability_SG}
     P(\tilde{y}_i = c|y_i = c) = \left\{\begin{array}{ll}
        SG(h_\varphi(x_i, y_i)), & c\neq k \\
        h_\varphi(x_i, y_i), & c=k \\
    \end{array}\right.
\end{equation}
where $SG$ means stop-gradient. Then the objective in Eq.(\ref{training_objective_1}) will be:
\begin{equation}
\label{training_objective_2}
    E[\log P(\tilde{\textbf{Y}}|\textbf{Y})] = \left\{\begin{array}{l}
		\displaystyle \sum_{i|\tilde{y}_i=k} \sum_{c\in \mathcal{C} \& c\neq k} SG(\log P(k|y_i=c)) * P(y_i=c) + \log P(k|y_i=k) * P(y_i=k) \\
		\displaystyle \sum_{i|\tilde{y}_i\neq k} \sum_{c \in \mathcal{C} \& c \neq k, c\neq \tilde{y}} SG(\log P(\tilde{y}_i|y_i=c)) *P(y_i=c) + \log(P(\tilde{y}_i|y_i=k)) P(y_i = k)
	\end{array}\right.
\end{equation}
In our formulation, for $|\mathcal{C}|$-class classification tasks, we have $2|\mathcal{C}|$ parameters to tune. However, in the implementation, we may shrink these parameters into several ones. Refer to the details in our implementation in the next section.

\subsection{Application in Recommendation tasks and Image Classification tasks}
\begin{algorithm}[h]
    \small
  % \SetKwInOut{Input}{Input}\SetKwInOut{Output}{Output}
  \KwIn{Corrupted data $\tilde{\textbf{Y}}$, learning rate $\beta$, epochs $T$, hyper-parameter $\alpha, C_1, C_2$, regularization weight $\lambda$, target recommender $f$, auxiliary model $g$, $h$, $h'$}
  \KwOut{Parameters $\theta, \mu, \phi, \psi$ for $f,g,h,h'$, correspondingly}
  Initialize all parameters\;
  $count$ $\leftarrow$ 0\; 
  \While{Not early stop and $epoch < T$}{
    Draw a minibatch of $(u,i_+)$ from $\{(u,i)| \tilde{y}_i = 1\}$\;
    Draw  $(u,i_-)$ from $\{(u,i)| \tilde{y}_i = 0\}$\ for each $(u,i_+)$\;
    % Sample one negative item for each user in the minibatch and constitute $k$ minibatches of $(u,i_-)$ from $\{(u,i)| \tilde{y}_i = 0\}$\;
    \uIf{$count \% 2 == 0$}{
      Compute $\mathcal{L}_{DeCA}$ according to Eq.(\ref{DPI-loss}) and Eq.(\ref{training_objective_1})\;
    }
    \uElse{
      Compute $\mathcal{L}_{DeCA}$ according to Eq.(\ref{DPI-loss}) and Eq.(\ref{training_objective_1})\;
    }
    Add regularization term: $\mathcal{L}_{DeCA} \leftarrow \mathcal{L}_{DeCA} + \lambda||\theta||^2$\;
    \For{each parameter $\Theta$ in $\{\theta, \mu, \phi, \psi\}$}{
      Compute $\partial \mathcal{L}_{DeCA} / \partial \Theta$ by back-propagation\;
      $\Theta \leftarrow \Theta - \beta \partial \mathcal{L}_{DeCA}/\partial \Theta$
    }
    $count$ $\leftarrow$ $count+1$\; 
  }
  \caption{Learning algorithm of DeCA for recommendation tasks}
  \label{alg:DPI}
\end{algorithm}

\begin{algorithm}[h]
\small
  \KwIn{Corrupted data $\tilde{\textbf{Y}}$, learning rate $\beta$, epochs $T$, hyper-parameter $\alpha, C_1, C_2$, regularization weight $\lambda$, target recommender $f$, auxiliary model $h$, $h'$}
  \KwOut{Parameters $\theta, \phi, \psi$ for $f,h,h'$, correspondingly}
  Set random seed to $s_1$, initialize another copy of $\theta$ as $\theta'$\;
  \While{Not early stop and $epoch < T$}{
    Draw a minibatch of $(u,i_+)$ from $\{(u,i)| \tilde{y}_i = 1\}$\;
    Draw  $(u,i_-)$ from $\{(u,i)| \tilde{y}_i = 0\}$\ for each $(u,i_+)$\;
    Compute binary cross-entropy 
    $\mathcal{L}_{BCE}$ on $(u,i_+)$ and $(u,i_-)$ with $f_{\theta'}$\;
    Add regularization: $\mathcal{L}_{BCE} \leftarrow \mathcal{L}_{BCE} + \lambda||\theta'||^2$\;
    Compute $\partial \mathcal{L}_{BCE}/\partial \theta'$ by back-propagation\;
    $\theta' \leftarrow \theta' - \beta \partial \mathcal{L}/\partial \theta'$\;
  }
  Freeze $\theta'$, set random seed to $s_2$ and initialize $\theta, \phi, \psi$\; 
  $count$ $\leftarrow$ 0\; 
  \While{Not early stop and $epoch < T$}{
    Draw $(u,i_+)$ and $(u,i_-)$ similarly with line3-4\;
    \uIf{$count \% 2 == 0$}{
      Compute $\mathcal{L}_{DeCA(p)}$ according to Eq.(\ref{DVAE-loss}) and Eq.(\ref{training_objective_1})\;
    }
    \uElse{
      Compute $\mathcal{L}_{DeCA(p)}$ according to Eq.(\ref{DVAE-loss}) and Eq.(\ref{training_objective_1})\;
    }
    Add regularization term: $\mathcal{L}_{DeCA(p)} \leftarrow \mathcal{L}_{DeCA(p)} + \lambda||\theta||^2$\;
    \For{each parameter $\Theta$ in $\{\theta, \phi, \psi\}$}{
      Compute $\partial \mathcal{L}_{DeCA(p)} / \partial \Theta$ by back-propagation\;
      $\Theta \leftarrow \Theta - \beta \partial \mathcal{L}_{DeCA(p)}/\partial \Theta$\;
    }
    $count$ $\leftarrow$ $count+1$\; 
  }
  \caption{Learning algorithm of DeCA(p) for recommendation tasks.}
  \label{alg:DVAE}
\end{algorithm}

\begin{algorithm}[h!]
    \KwIn{Corrupted data $\tilde{\textbf{Y}}$, learning rate $\beta$, number of epochs $T_1, T_2$, hyper-parameter $\alpha, C_1, C_2$, regularization weight $\lambda$, target model $f$, auxiliary model $h$}
  \KwOut{Parameters $\theta, \phi, \psi$ for $f,h,h'$, correspondingly}
  Set random seed to $s_1$, initialize another copy of $\theta$ as $\theta'$\;
  \While{Not early stop and $epoch < (T_1 + T_2)$}{
    Draw a minibatch $(x, y)$ from $\textbf{X}$ and $\textbf{Y}$, Computer binary cross-entropy loss: $\mathcal{L}_{cross-entropy}(f_{\theta'}(x), y)$\;
    Compute $\partial \mathcal{L}_{cross-entropy}/\partial \theta'$ by back-propagation\;
    $\theta' \leftarrow \theta' - \beta \partial \mathcal{L}_{cross-entropy}/\partial \theta'$\;
  }
  Freeze $\theta'$, set random seed to $s_2$ ($s_2 \neq s_1$) and initialize $\theta, \phi, \psi$\; 
  $count$ $\leftarrow$ 0\; 
  \While{Not early stop and $epoch < (T_1+T_2)$}{
    Draw a minibatch $(x, y)$ from $\textbf{X}$ and $\textbf{Y}$\;
    $k \leftarrow count \% |\mathcal{C}|$\;
    \uIf{$epoch <= T_1$}{
        Computer $\mathcal{L}$ according to Eq.(\ref{training_objective_1})\;
    }
    \uElse{
        Computer $\mathcal{L}$ according to Eq.(\ref{training_objective_2})\;
    }
    \For{each parameter $\Theta$ in $\{\theta, \phi, \psi\}$}{
      Compute $\partial \mathcal{L}_{DeCA(p)} / \partial \Theta$ by back-propagation\;
      $\Theta \leftarrow \Theta - \beta \partial \mathcal{L}_{DeCA(p)}/\partial \Theta$\;
    }
    $count$ $\leftarrow$ $count+1$\; 
  }
    \caption{Learning algorithm of DeCA(p) for multi-class classification problems.}
    \label{alg:multiclass-DeCAp}
\end{algorithm}

In this section, we introduce how to apply the above proposed methods into recommendation tasks (binary-classification problems) and image classification tasks (multi-class classification problems), in order to show the effectiveness of our methods. 

For recommendation tasks, the input $x_i$ is actually a user-item pair. So we simply need to replace the $x_i$ with $(u,i)$ where $u,i$ means the pair of user $u$ and item $i$. Then we give the pseudo code of DeCA and DeCA(p) on recommendation tasks in Algorithm \ref{alg:DPI} and Algorithm \ref{alg:DVAE}, respectively. Note that the noises in the positive and negative samples have different sources. Thus when focusing on different classes, $C_k$ should be different. ($C_{k1}$ and $C_{k2}$ in Eq.(\ref{training_objective_1}) shrink to $C_k$ in binary tasks, see Section \ref{ssec:training_routine}). Thus in our experiments, we tuned two parameters $C_0$ and $C_1$, and the results also show that setting $C_0$ and $C_1$ to the same numbers is not the optimal solution. 
\wyy{
In our training process, there are two classes $\mathcal{C} = \{0, 1\}$. Then when $k=0$,
$-\log P(\tilde{y}_i=k|y_i \neq k) = -\log P(\tilde{y}_i=0|y_i=1) $ is modelled by $h_\psi'$. 
In the meantime, $-\log P(\tilde{y}_i=1|y_i=0)$ will be replaced with $C_0$. 
In this case, we are actually assuming that $P(\tilde{y}_i=0|y_i=0) = 1$, thus under this assumption,
for all the samples with $\tilde{y}_i=1$, the corresponding $y_i$ could only be $1$, 
or there would be conflicts. 
So when $k=0$, we are denoising negative samples. 
And we call this step \textbf{DN} (Denoising Negative). Similarly, when $k=1$, the corresponding step is called \textbf{DP} (Denoising Positive). 
\noindent Note that we do not need to calculate the equation as in Eq.(\ref{training_objective_2}) since in our experiments training with Eq.(\ref{training_objective_2}) does not bring any improvements. We remove this part to make the process more simplified. 
%  For image classification tasks, this strategy could sometimes yield slightly better results. The reason for this could be that compared to Eq.(\ref{training_routine_set_conditional_probability}), Eq.(\ref{training_routine_set_conditional_probability}) is smoother on the classes $c\neq k$. When $h_\varphi$ is well-learnt, the latter could lead to better performances especially in the cases that $h_\varphi(x_i, y_i)$ is much more precise than approximating the probability with $1$. 
}

\noindent Besides, for Co-trained DeCA, we chose $g_\mu$ to be MF to (a)  achieve more efficient computation and more stable learning because of the simplicity of MF; (b) yield good performances since MF is shown to be still effective (as stated in Section \ref{sub:notations_and_problem_formulation}); (c) make the proposed DeCA more applicable without tuning on the selection of $g_\mu$, as MF is one of the most general models. While for $f_\theta$, the target model, it could be any recommendation model. As for $h_\phi$ and $h_\psi'$ in Eq.(\ref{h_model}), we also implement them as MF. In ablation study, we try some other models for $h_\phi$ and $h_\psi'$ but find no differences. 

For image classification tasks, the input $x_i$ is the image. we also give the pseudo code of DeCA(p) on this task in Algorithm \ref{alg:multiclass-DeCAp}. In our experiments, we focus on random noises in the labels. \wyy{Then since the sources of noises in different classes are the same, we intuitively set $C_{01} = C_{11} = \cdots = C_{|\mathcal{C}|1}$ and $C_{02} = C_{12} = \cdots = C_{|\mathcal{C}|2}$. Then we denote $C_1 = C_{01}$, $C_2 = C_{02}$. 
When focusing on the class $k$, we want to highlight every value corresponding with class $k$, so we do not merge $C_1$ and $C_2$ to the same value. Instead, we tune two parameters $C_1$ and $C_2$ in our experiments. }

The values of $C_{k1}$ and $C_{k2}$, $k\in \mathcal{C}$ are the key to handle different kinds of noises. When applying DeCA into different problems, we need to think about the inner property of the problem and carefully choose the values and then tune them. 

In our implementation, $f_\theta$ serves as the image classification model, and $h_\varphi$ is a two-layer MLP with ReLU activation. To reduce the number of parameters to be trained and the burden of training, we opt to feed the embeddings of the images from the last but one layer of $f_\theta$ into $h_\varphi$ instead of reading images directly.}

\subsection{Discussion}

In this subsection, we provide a brief discussion to illustrate the relationship between our methods and other related methods.  
\subsubsection{Relationship with Variational Auto-Encoder}
We can see that Eq.(\ref{VAE_lower_bound}) is exactly the objective function of a variational auto-encoder (VAE) \cite{doersch2016vaetutorial}.
Specifically, the real user preference $\textbf{Y}$ is the latent variable. $P_f(\textbf{Y}|\tilde{\textbf{Y}})$ which is parameterized by our target model $f_\theta$ maps the corrupted data $\tilde{\textbf{Y}}$ to the latent variables $\textbf{Y}$ and thus can be seen as the encoder. The likelihood $P(\tilde{\textbf{Y}}|\textbf{Y})$ describes the distribution of corrupted data $\tilde{\textbf{Y}}$ given the real user preference $\textbf{Y}$ and acts as the decoder. Finally, $P(\mathbf{R})$ is the fixed prior distribution. However, we claim that the proposed DeCA(p) is substantially different from VAE. For example, DeCA(p) does not utilize the re-parameterization of VAE. VAE is just an interpretation of DeCA(p). Besides, although there are some methods utilizing VAE for recommendation ~\cite{DBLP:conf/www/LiangKHJ18,DBLP:conf/wsdm/ShenbinATMN20}, few of them are designed for recommendation denoising. Finally, existing VAE-based recommendation methods are actually recommendation models while the proposed DeCA(p) acts a \emph{model-agnostic training framework} which is then instantiated with all kinds of downstream models. They serve on different levels.  

% \subsubsection{Analysis of the Convergence}
% During training, if the parameters $C_1$ and $C_2$ are large enough, the process will act similarly to the normal training, since (1) the gradients from the fixed part will dominate the update of the model when $C_1$ and $C_2$ are large. (2) With large $C_1$ and $C_2$, we demonstrate that the objective has the same target as cross-entropy loss as follows. 

% If we simply replace $-\log(1-h'_\psi(u, i))$ and $-\log(1-h_\phi(u, i))$ with $C_1$ and $C_2$ in the term $E[\log P(\tilde{\textbf{Y}}|\textbf{Y})]$, we could obtain the following expansion: 
% \begin{equation}\label{eq:expansion}
%     E[\log P(\tilde{\textbf{Y}}|\textbf{Y})] = - \sum_{(u,i)|\tilde{y}_i = 0} C_1 \cdot f_\theta(u,i) - \sum_{(u,i)|\tilde{y}_i = 1} C_2 \cdot (1 - f_\theta(u,i))
% \end{equation}
% Thus maximizing Eq.(\ref{eq:expansion}) will optimize $f_\theta(u,i)$ to fit the observational data. The loss for the dataset is similar to cross-entropy loss, and it will converge. 

\subsubsection{Analysis of the Run Time Complexity}
For DeCA, the main additional complexity comes from the forward and backward of the auxiliary model. DeCA also incorporates information from different models. However, compared with ensemble methods, DeCA only uses the target model for inference while ensemble methods need the forward pass of each model. As a result, DeCA is more efficient than ensemble methods in the inference stage.

% $g_\mu$. However, since MF is the simplest model, the time complexity will be much smaller than twice the complexity of normal training. 

% For DeCA(p), the total time complexity will be almost exactly twice of normal training, since  the complexity of the second time is almost the same as the first time, i.e., normal training.

\section{EXPERIMENTS} % (fold)
\label{sec:experiments}

In this section, we empirically demonstrate the effectiveness of the proposed DeCA through extensive experiments. For the binary scenario, we instantiate the proposed DeCA and DeCA(p) with four state-of-the-art recommendation models as the target model $f$. For the multi-class scenario , we instantiate DeCA(p) with the ResNet32 model as the target model $f$. We aim to answer the following research questions:
% \href{}{GoogleDrive}
\begin{itemize}[leftmargin=*]
  \item \textbf{RQ1}: How do the proposed methods perform compared to normal training and other denoising methods? Can the proposed methods help to downgrade the effect of noisy examples?
  \item \textbf{RQ2}: How does the design of proposed methods affect the recommendation performance, including the iterative training, hyperparameter study, and model selection?
  \item \textbf{RQ3}: Can the proposed methods generate reasonable preference distribution given the corrupted binary recommendation data?
%   \item \textbf{RQ4}: How do the proposed methods perform for the multi-class scenario?
\end{itemize}
\subsection{Experimental Settings} % (fold)
\label{sub:experimental_settings}
\subsubsection{Datasets} % (fold)
\label{ssub:datasets}
For recommendation tasks, we conduct experiments with four public accessible datasets: MovieLens, Modcloth\footnote{https://github.com/MengtingWan/marketBias}, Adressa\footnote{https://github.com/WenjieWWJ/DenoisingRec} and Electronics\footnote{https://github.com/MengtingWan/marketBias}. 
More statistics about these three datasets are listed in Table \ref{tab:dataset statistics}. For each dataset, we construct the clean test set with only clean examples which denote the real user preference. 

\begin{table}[t]
\begin{center}
\caption{Statistics of the datasets}
% \vspace{-7pt}
\label{tab:dataset statistics}
\begin{tabular}{c|c|c|c|c}
  \toprule
  Dataset & \# Users & \# Items & \# Interactions & Sparsity \\
  \midrule
  MovieLens & 943 & 1,682 & 100,000 & 0.93695 \\
%   \hline
  Modcloth & 44,783 & 1,020 & 99,893 & 0.99781 \\
%   \hline 
  Adressa & 212,231 & 6,596 & 419,491 & 0.99970 \\
%   \hline
  Electronics & 1,157,633 & 9,560 & 1,292,954 & 0.99988 \\
  \bottomrule
\end{tabular}
\end{center}
% \vspace{-0.5cm}
\end{table}

\textbf{MovieLens}~\cite{DBLP:journals/tiis/HarperK16}. This is one of the most popular datasets in the task of recommendation. For evaluation, the clean test set is constructed based on user-item pairs with ratings equal to 5. 

\textbf{Modcloth}. It is from an e-commerce website that sells women's clothing and accessories. For evaluation, the clean test set is built on user-item pairs whose rating scores are equal to 5.

\textbf{Adressa}. This is a real-world news reading dataset from Adressavisen. It contains the interaction records of anonymous users and the news. %The dwell time for each user-item pair is used to construct the clean test set. 
According to \cite{kim2014modeling}, we use interactions with dwell time longer than 10 seconds to construct the clean test set.

\textbf{Electronics}. Electronics is collected from the \emph{Electronics} category on Amazon \cite{ni2019justifying,wan2020addressing}. %It is based on the public \emph{Amazon 2018 Dataset}\cite{ni2019justifying} and further processed by \cite{wan2020addressing}. 
The clean test set is built with user-item pairs with rating scores equal to 5.

Note that all the ratings and dwell time are only used to construct the clean test set. The recommendation models are trained with only the corrupted binary implicit feedback.

\textbf{CIFAR10}. It consists of 60000 32x32 colour images in 10 classes, with 6000 images per class. There are 50000 training images and 10000 test images.
% \item \textbf{CIFAR5}. It is the dataset sampled from CIFAR10 which contains the data corresponding to the classes \{0, 1, 2, 3, 4\} in CIFAR10. 

\textbf{Fashion-MNIST}\cite{xiao2017/online}. It is a dataset of Zalando's article images—consisting of a training set of 60,000 examples and a test set of 10,000 examples. Each example is a 28x28 grayscale image, associated with a label from 10 classes.

\subsubsection{Evaluation protocols} % (fold)
\label{ssub:evaluation_protocals}
We adopt cross-validation to evaluate the performance. 
For Adressa and MovieLens, we split the user-item interactions into the training set, validation set, and test set according to the ratio of 8:1:1 in chronological order \cite{wang2020denoising}. As for Modcloth and Electronics, we randomly split the historical interactions according to the ratio of 8:1:1. After that, the clean test set is constructed for each dataset as described in section \ref{ssub:datasets}. For CIFAR10 and Fashion-MNIST, we use the default split, where 50000 pictures are the training set and 10000 pictures are the testing set. Then we randomly sample 10000 pictures from the training set to construct the validation set. 

For recommendation tasks, the performance is measured by two widely used top-$K$ recommendation metrics \cite{he2017neural, yang2018hop}: recall@$K$ and ndcg@$K$. By default, we set K = 5, 20 for Modcloth and Adressa, and K = 10, 50 for Electronics since the number of items in Electronics is larger. The metric accuracy is used for the image classification tasks. All experiments are run 3 times. The average and standard deviation are reported. 

\subsubsection{Baselines} % (fold)
\label{ssub:baselines}
\wy{For recommendation tasks,} We select four state-of-the-art recommendation models as the target model $f$ of DeCA and DeCA(p):
\begin{itemize}[leftmargin=*]
  \item \textbf{GMF}~\cite{he2017neural}: This is a generalized version of MF by changing the inner product to the element-wise product and a dense layer.
  \item \textbf{NeuMF}~\cite{he2017neural}: The method is a state-of-the-art neural CF model which combines GMF with a Multi-Layer Perceptron (MLP).
  \item \textbf{CDAE}~\cite{wu2016collaborative}: CDAE corrupts the 
  observed interactions %%% 
  with random noises, and then employs several linear layers to reconstruct the original datasets, which will increase its anti-noise abilities.
  \item \textbf{LightGCN}~\cite{he2020lightgcn}: LightGCN is a newly proposed graph-based recommendation model which learns user and item embeddings by linearly propagating them on the interaction graph.% and uses the weighted sum of the embeddings learned at all layers as the final embedding, which is easy to implement and train, yet with excellent performance.
\end{itemize}
Each model is trained with the following approaches:
\begin{itemize}[leftmargin=*]
  \item \textbf{Normal}: Train the model on the noisy dataset with simple binary-cross-entropy (BCE) loss.
  \item \textbf{WBPR}~\cite{gantner2012personalized}: This is a re-sampling based denoising method, which considers the popular but uninteracted items are highly likely to be real negative ones.
  \item \textbf{T-CE}~\cite{wang2020denoising}: This is a re-weighting based denoising method, which uses the Truncated BCE to assign zero weights to large-loss examples with a dynamic threshold in each iteration.
  \item \textbf{Ensemble}: This is an ensemble-based approach, which aggregates the results from two models with different random seeds.
%   examples with high loss values 
  \item \textbf{DeCA} and \textbf{DeCA(p)}: Our proposed methods.
\end{itemize}
Besides, \cite{yu2020sampler} proposed a noisy robust learning method. We do not compare with this method because it has been shown to be only applicable to the MF model \cite{yu2020sampler}.

For image classification tasks, we choose ResNet-32~\cite{resnet} as the backbone model. Besides, we select the following denoising methods as the baselines: 
\begin{itemize}[leftmargin=*]
    \item \textbf{Normal}: Training on the noisy dataset. 
    \item \textbf{MW-Net}~\cite{shu2019meta}: It build a meta-weight-net to reweight the loss for each instance. The meta-weight-net is trained using a small set of clean data. In our implementaion, we select 1000 images in the whole dataset before adding noise to the remaining part, which is consistent with \cite{shu2019meta}. This serves as the re-weighting strategy mentioned in Section \ref{introduction}.
    \item \textbf{ITLM}~\cite{shen2019learning}: It uses the iterative trimmed loss strategy, select a certain portion of the dataset (\emph{eg.} 80\% of the dataset) with the minimal loss to construct a new subset, which will be used to train the model in next iteration. This serves as the re-sampling strategy mentioned in Sectioin \ref{introduction}. 
    \item \textbf{DeCA(p)}: Our proposed method.
\end{itemize}

\subsubsection{Parameter settings} % (fold)
\label{ssub:parameter_settings}
We optimize all models using Adam optimizer. For recommendation tasks, The batch size is set as 2048. The learning rate is tuned as 0.001 on four datasets.
For each training instance, we sample one interacted sample and one randomly sampled missing interaction to feed the model. 
We use the recommended network settings of all models.
Besides, the embedding size of users and items in GMF and NeuMF, the hidden size of CDAE, are all set to 32. 
For LightGCN, the embedding size is set to 64 without dropout \cite{he2020lightgcn}. The $L_2$ regularization coefficient is tuned in $\{0.01, 0.1, 1, 10, 100, 1000\}$ divided by the number of users in each dataset.
For DeCA and DeCA(p), there are three hyper-parameters: $C_1$, $C_2$ and $\alpha$. 
We apply a grid search for hyperparameters: $C_1$ and $C_2$ are searched among \{1, 10, 100, 1000\},  $\alpha$ is tuned in \{0, 0.5, 1\}. 
For image classification tasks, the batch size is set to 100, and the learning rate is tuned in [0.1, 0.01]. The hyperpameters $C_1$ and $C_2$ are searched among \{1, 10, 50, 100\}.
Note that the hyperparameters of models (\emph{e.g.} the hidden size of the model) keep exactly the same across all training approaches for a fair comparison. It also indicates that the proposed methods can be easily integrated with downstream recommendation models without exhaustive hyperparameter refinement.

\subsection{Performance Comparison (RQ1)} % (fold)
\label{sub:performance_comparison}

\subsubsection{Recommendation}
Table \ref{tab:Overall_performance_on_movielens_and_modcloth} and Table \ref{tab:Overall_performance_on_adressa_and_electronics} shows the performance comparison on Movielens, Modcloth, Adressa and Electronics.
From the table, we can have the following observations:
\begin{itemize}[leftmargin=*]
  \item The proposed DeCA and DeCA(p) can effectively improve the recommendation performance of all the four recommendation models over all datasets. Either DeCA or DeCA(p) achieves the best performance compared with normal training and other denoising methods, except for few cases. Even if the CDAE model itself is based on the denoising auto-encoder which is more robust to corrupted data, there still exists improvement when training with our proposed methods, especially in the MovieLens dataset.
  \item Simple ensemble of the results from different models sometimes cannot lead to better performance, for example, in the Electronic dataset. The reason could be that the Electronic data is super sparse. The results from different models vary much more. Simple aggregating two very different results could lead to poor performance in such cases.  
  \item The performance of LightGCN is not so good on Modcloth, Adressa and Electronics. This could be attributed to the bias of datasets.
  We can see from Table \ref{tab:dataset statistics} that users in three datasets would only be connected to a small number of items while items are connected to a large amount of users. The imbalance of interaction graphs could affect the performance of LightGCN.
\end{itemize}

\begin{table*}[ht]
\centering
\footnotesize
\caption{Overall performance comparison on MovieLens and Modcloth. The highest scores are in Boldface. R is short for Recall and N is short for NDCG. The results with improvements over the best baseline larger than 5\% are marked with $*$.}
\vspace{-0.3cm}
\label{tab:Overall_performance_on_movielens_and_modcloth}
\resizebox{\textwidth}{!}{%
\begin{tabular}{c|cccc|cccc}
\toprule
\multirow{2}{*}{\textbf{MovieLens}}& R@3 & R@20 &N@3 &N@20 & R@3 & R@20 &N@3 &N@20 \\
\cmidrule(lr){2-5}\cmidrule{6-9}
 &\multicolumn{4}{c|}{GMF} & \multicolumn{4}{c}{NeuMF} \\ 
    \midrule
Normal & 0.021$\pm$0.001 & 0.095$\pm$0.001 & 0.035$\pm$0.003 & 0.059$\pm$0.001 & 0.025$\pm$0.002 & 0.103$\pm$0.006 & 0.041$\pm$0.002 & 0.064$\pm$0.003 \\  
WBPR & 0.023$\pm$0.002 & 0.082$\pm$0.003 & 0.045$\pm$0.002 & 0.056$\pm$0.000 & 0.022$\pm$0.003 & 0.086$\pm$0.001 & 0.038$\pm$0.005 & 0.054$\pm$0.002 \\ 
T-CE & 0.017$\pm$0.002 & 0.098$\pm$0.001 & 0.026$\pm$0.003 & 0.054$\pm$0.001 &  0.026$\pm$0.004 & 0.106$\pm$0.002 & 0.047$\pm$0.003 & 0.068$\pm$0.002 \\ 
Ensemble & 0.021$\pm$0.000 & 0.095$\pm$0.001 & 0.035$\pm$0.001 & 0.059$\pm$0.000 & 0.030$\pm$0.001 & 0.108$\pm$0.003 & 0.048$\pm$0.002 & 0.070$\pm$0.002 \\ 
DeCA & 0.022$\pm$0.001 & \textbf{0.109$\pm$0.007$^*$}
& 0.038$\pm$0.001 & 0.064$\pm$0.002 & 0.022$\pm$0.001 & 0.099$\pm$0.006 & 0.032$\pm$0.003 & 0.059$\pm$0.003 \\ 
DeCA(p) & \textbf{0.027$\pm$0.002$^*$} & 0.099$\pm$0.001 & \textbf{0.054$\pm$0.003$^*$} & \textbf{0.066$\pm$0.001$^*$} & \textbf{0.035$\pm$0.000$^*$} & \textbf{0.120$\pm$0.003$^*$} & \textbf{0.064$\pm$0.000$^*$} & \textbf{0.081$\pm$0.002$^*$} \\ 
\midrule
& \multicolumn{4}{c|}{CDAE} & \multicolumn{4}{c}{LightGCN} \\ 
\midrule
Normal & 0.017$\pm$0.000 & 0.095$\pm$0.002 & 0.028$\pm$0.004 & 0.052$\pm$0.001 & 0.025$\pm$0.000 & 0.106$\pm$0.001 & 0.048$\pm$0.000 & 0.065$\pm$0.000 \\
WBPR & 0.017$\pm$0.003 & 0.087$\pm$0.004 & 0.028$\pm$0.003 & 0.050$\pm$0.002 & \textbf{0.031}$\pm$0.000 & 0.093$\pm$0.001 & 0.054$\pm$0.000 & 0.064$\pm$0.000 \\ 
T-CE & 0.014$\pm$0.002 & 0.095$\pm$0.007 & 0.024$\pm$0.004 & 0.050$\pm$0.001 & 0.011$\pm$0.001 & 0.080$\pm$0.007 & 0.020$\pm$0.002 & 0.043$\pm$0.002 \\  
Ensemble & 0.015$\pm$0.003 & 0.099$\pm$0.000 & 0.026$\pm$0.004 & 0.052$\pm$0.001 & 0.026$\pm$0.001 & 0.105$\pm$0.002 & 0.049$\pm$0.001 & 0.064$\pm$0.000 \\ 
DeCA & \textbf{0.029$\pm$0.001$^*$} & 0.109$\pm$0.000 & \textbf{0.048$\pm$0.001$^*$} & 0.066$\pm$0.000 & 0.027$\pm$0.000 & 0.104$\pm$0.001 & 0.044$\pm$0.002 & 0.064$\pm$0.001 \\  
DeCA(p) & 0.028$\pm$0.002 & \textbf{0.110$\pm$0.002$^*$} & 0.047$\pm$0.002 & \textbf{0.067$\pm$0.001$^*$} & 0.029$\pm$0.001 & \textbf{0.118$\pm$0.001$^*$} & \textbf{0.055}$\pm$0.001 & \textbf{0.075$\pm$0.001$^*$} \\ 
\bottomrule
\toprule
\multirow{2}{*}{\textbf{Modcloth}}& R@5 & R@20 &N@5 &N@20 & R@5 & R@20 &N@5 &N@20 \\
\cmidrule(lr){2-5}\cmidrule{6-9}
 &\multicolumn{4}{c|}{GMF} & \multicolumn{4}{c}{NeuMF} \\ 
    \hline
Normal & 0.063$\pm$0.006 & 0.225$\pm$0.008 & 0.043$\pm$0.005 & 0.088$\pm$0.005 & 0.082$\pm$0.003 & 0.242$\pm$0.004 & 0.055$\pm$0.002 & 0.101$\pm$0.002 \\
WBPR & 0.067$\pm$0.002 & 0.224$\pm$0.002 & 0.046$\pm$0.001 & 0.090$\pm$0.001 & 0.092$\pm$0.001 & 0.247$\pm$0.020 & 0.064$\pm$0.001 & 0.108$\pm$0.006 \\
T-CE & 0.067$\pm$0.004 & 0.235$\pm$0.002 & 0.045$\pm$0.003 & 0.093$\pm$0.001 & 0.065$\pm$0.010 & 0.228$\pm$0.008 & 0.044$\pm$0.010 & 0.091$\pm$0.008 \\
Ensemble & 0.064$\pm$0.004 & 0.228$\pm$0.007 & 0.043$\pm$0.002 & 0.089$\pm$0.002 & 0.090$\pm$0.003 & 0.260$\pm$0.004 & 0.061$\pm$0.001 & 0.109$\pm$0.002 \\ 
DeCA & 0.072$\pm$0.001 & 0.245$\pm$0.001 & 0.050$\pm$0.001 & 0.099$\pm$0.001 & \textbf{0.099$\pm$0.005$^*$} & \textbf{0.268}$\pm$0.006 & \textbf{0.065}$\pm$0.001 & \textbf{0.113}$\pm$0.001 \\
DeCA(p) & \textbf{0.074$\pm$0.001$^*$} & \textbf{0.247$\pm$0.001$^*$} & \textbf{0.052$\pm$0.001$^*$} & \textbf{0.100$\pm$0.000$^*$} & 0.087$\pm$0.005 & 0.265$\pm$0.006 & 0.059$\pm$0.004 & 0.110$\pm$0.004 \\
\midrule
& \multicolumn{4}{c|}{CDAE} & \multicolumn{4}{c}{LightGCN} \\ 
\midrule
Normal & 0.082$\pm$0.004 & 0.242$\pm$0.003 & 0.052$\pm$0.002 & 0.098$\pm$0.001 & 0.065$\pm$0.001 & 0.220$\pm$0.002 & 0.043$\pm$0.001 & 0.087$\pm$0.002 \\
WBPR & 0.079$\pm$0.000 & 0.238$\pm$0.004 & 0.050$\pm$0.000 & 0.095$\pm$0.001 & 0.072$\pm$0.001 & 0.222$\pm$0.002 & 0.046$\pm$0.001 & 0.088$\pm$0.001 \\
T-CE & 0.075$\pm$0.003 & 0.243$\pm$0.008 & 0.048$\pm$0.002 & 0.096$\pm$0.002 & 0.071$\pm$0.000 & 0.231$\pm$0.001 & 0.049$\pm$0.000 & 0.093$\pm$0.000 \\
Ensemble & 0.084$\pm$0.002 & 0.250$\pm$0.002 & 0.054$\pm$0.001 & 0.100$\pm$0.001 & 0.068$\pm$0.001 & 0.227$\pm$0.001 & 0.046$\pm$0.000 & 0.090$\pm$0.000 \\ 
DeCA & 0.086$\pm$0.004 & 0.250$\pm$0.003 & 0.056$\pm$0.001 & 0.102$\pm$0.000 & 0.064$\pm$0.001 & 0.221$\pm$0.002 & 0.041$\pm$0.000 & 0.085$\pm$0.000 \\
DeCA(p) & \textbf{0.089$\pm$0.004$^*$} & \textbf{0.251}$\pm$0.005 & \textbf{0.057$\pm$0.003$^*$} & \textbf{0.103$\pm$0.003$^*$} & \textbf{0.073$\pm$0.001} & \textbf{0.235$\pm$0.001} & \textbf{0.051$\pm$0.001} & \textbf{0.096$\pm$0.000} \\
\bottomrule
\end{tabular}}
\vspace{-0.2cm}
% \vspace{-5pt}
\end{table*}

\begin{table*}[ht]
\centering
\footnotesize
\caption{Overall performance comparison on Adressa and Electronics. The highest scores are in Boldface. R is short for Recall and N is short for NDCG. The results with improvements over the best baseline larger than 5\% are marked with $*$.}
\vspace{-0.3cm}
\label{tab:Overall_performance_on_adressa_and_electronics}
\resizebox{\textwidth}{!}{%
\begin{tabular}{c|cccc|cccc}
\toprule
\multirow{2}{*}{
    \textbf{Adressa}} & R@5 & R@20 &N@5 &N@20 & R@5 & R@20 &N@5 &N@20 \\
\cmidrule(lr){2-5}\cmidrule{6-9} & \multicolumn{4}{c|}{GMF} & \multicolumn{4}{c}{NeuMF} \\
    \hline
Normal&0.116$\pm$0.003&0.209$\pm$0.005&0.080$\pm$0.002&0.112$\pm$0.001&0.169$\pm$0.004&0.312$\pm$0.004&0.131$\pm$0.002&0.180$\pm$0.003 \\
WBPR&0.115$\pm$0.007&0.210$\pm$0.007&0.084$\pm$0.003&0.116$\pm$0.003&0.172$\pm$0.002&0.311$\pm$0.003&0.132$\pm$0.001&0.181$\pm$0.001 \\
T-CE&0.109$\pm$0.000&0.209$\pm$0.001&0.070$\pm$0.000&0.104$\pm$0.000&0.172$\pm$0.003&0.312$\pm$0.003&0.134$\pm$0.001&\textbf{0.183$\pm$0.002} \\
Ensemble & 0.110$\pm$0.002 & 0.191$\pm$0.006 & 0.078$\pm$0.002 & 0.105$\pm$0.003 & 0.180$\pm$0.001 & 0.311$\pm$0.001 & 0.134$\pm$0.001 & 0.180$\pm$0.000   \\
DeCA&\textbf{0.125$\pm$0.002$^*$} &\textbf{0.220$\pm$0.001$^*$} & 0.091$\pm$0.002 & 0.126$\pm$0.002 & 0.170$\pm$0.006 & \textbf{0.318$\pm$0.002} & 0.130$\pm$0.001 & 0.181$\pm$0.000 \\
DeCA(p)& 0.123$\pm$0.005 &  0.220$\pm$0.001 & \textbf{0.093$\pm$0.004$^*$} & \textbf{0.127$\pm$0.003$^*$} & \textbf{0.183$\pm$0.009}&0.316$\pm$0.004&\textbf{0.137$\pm$0.002}&0.181$\pm$0.005 \\
\midrule
  & \multicolumn{4}{c|}{CDAE} & \multicolumn{4}{c}{LightGCN} \\ 
\midrule
Normal & 0.162$\pm$0.000 & 0.317$\pm$0.001 & 0.123$\pm$0.000 & 0.178$\pm$0.000 & 0.085$\pm$0.004 & 0.215$\pm$0.005 & 0.064$\pm$0.003 & 0.107$\pm$0.003 \\
WBPR & 0.161$\pm$0.001 & 0.315$\pm$0.005 & 0.121$\pm$0.002 & 0.173$\pm$0.004 & 0.118$\pm$0.003 & 0.211$\pm$0.006 & 0.089$\pm$0.002 & 0.119$\pm$0.004 \\
T-CE & 0.161$\pm$0.001 & 0.317$\pm$0.003 & 0.122$\pm$0.002 & 0.176$\pm$0.004 & 0.119$\pm$0.001 & 0.206$\pm$0.003 & \textbf{0.091}$\pm$0.001 & 0.121$\pm$0.001 \\
Ensemble &  0.162$\pm$0.000 & 0.317$\pm$0.001 & 0.122$\pm$0.000 & 0.176$\pm$0.000 & 0.104$\pm$0.001 & 0.217$\pm$0.001 & 0.078$\pm$0.001 & 0.118$\pm$0.001 \\
DeCA & 0.162$\pm$0.000 & 0.319$\pm$0.000 & 0.123$\pm$0.000 & 0.178$\pm$0.000 & 0.112$\pm$0.001 & 0.221$\pm$0.001 & 0.077$\pm$0.005 & 0.116$\pm$0.005 \\
DeCA(p) & \textbf{0.163$\pm$0.000} & \textbf{0.320$\pm$0.002} & \textbf{0.123$\pm$0.000} & \textbf{0.178$\pm$0.001} & \textbf{0.121$\pm$0.001} & \textbf{0.222$\pm$0.001} & 0.089$\pm$0.001 & \textbf{0.125$\pm$0.000} \\
\bottomrule
\toprule
 \multirow{2}{*}{\textbf{Electronics}} & R@10 & R@50 &N@10 &N@50 & R@10 & R@50 &N@10 &N@50 \\
\cmidrule(lr){2-5}\cmidrule{6-9} 
 & \multicolumn{4}{c|}{GMF} & \multicolumn{4}{c}{NeuMF} \\ 
\midrule
Normal & 0.023$\pm$0.000 & 0.063$\pm$0.001 & 0.013$\pm$0.000 & 0.021$\pm$0.000 & 0.075$\pm$0.000 & 0.191$\pm$0.002 & 0.048$\pm$0.000 & 0.072$\pm$0.001 \\
WBPR & 0.026$\pm$0.001 & 0.072$\pm$0.001 & 0.014$\pm$0.000 & 0.024$\pm$0.000 & \textbf{0.077}$\pm$0.000 & \textbf{0.201}$\pm$0.001 & \textbf{0.048}$\pm$0.000 & 0.074$\pm$0.000 \\
T-CE & 0.024$\pm$0.000 & 0.063$\pm$0.001 & 0.013$\pm$0.000 & 0.022$\pm$0.000 & 0.071$\pm$0.004 & 0.189$\pm$0.008 & 0.045$\pm$0.002 & 0.071$\pm$0.003 \\
Ensemble & 0.014$\pm$0.001 & 0.034$\pm$0.001 & 0.010$\pm$0.000 & 0.016$\pm$0.000 &  0.051$\pm$0.000 & 0.108$\pm$0.001 & 0.040$\pm$0.000 & 0.055$\pm$0.000 \\
DeCA & \textbf{0.045$\pm$0.001$^*$} & 0.104$\pm$0.000 & 0.023$\pm$0.000 & 0.037$\pm$0.000 & 0.071$\pm$0.001 & 0.191$\pm$0.001 & 0.044$\pm$0.002 & 0.069$\pm$0.001 \\
DeCA(p) & 0.042$\pm$0.001 & \textbf{0.110$\pm$0.000$^*$} & \textbf{0.024$\pm$0.000$^*$} & \textbf{0.038$\pm$0.000$^*$} & 0.074$\pm$0.002 & 0.200$\pm$0.001 & 0.047$\pm$0.000 & \textbf{0.074}$\pm$0.000 \\
\midrule
& \multicolumn{4}{c|}{CDAE} & \multicolumn{4}{c}{LightGCN} \\ 
\midrule
Normal & 0.074$\pm$0.001 & 0.197$\pm$0.001 & 0.047$\pm$0.000 & 0.073$\pm$0.000 & 0.024$\pm$0.000 & 0.081$\pm$0.001 & 0.016$\pm$0.000 & 0.030$\pm$0.000 \\
WBPR & 0.077$\pm$0.000 & 0.190$\pm$0.000 & 0.048$\pm$0.000 & 0.072$\pm$0.000 & 0.022$\pm$0.000 & 0.076$\pm$0.001 & 0.015$\pm$0.000 & 0.028$\pm$0.000 \\
T-CE & 0.067$\pm$0.002 & 0.187$\pm$0.010 & 0.044$\pm$0.001 & 0.067$\pm$0.003 & 0.024$\pm$0.001 & 0.086$\pm$0.001 & 0.017$\pm$0.000 & 0.031$\pm$0.000 \\
Ensemble & 0.043$\pm$0.000 & 0.110$\pm$0.001 & 0.036$\pm$0.000 & 0.056$\pm$0.000 & 0.018$\pm$0.000 & 0.049$\pm$0.001 & 0.014$\pm$0.000 & 0.023$\pm$0.000 \\
DeCA & \textbf{0.077}$\pm$0.000 & 0.201$\pm$0.001 & 0.048$\pm$0.000 & 0.075$\pm$0.000 & 0.025$\pm$0.000 & 0.103$\pm$0.001 & 0.018$\pm$0.000 & 0.036$\pm$0.000 \\
DeCA(p) & 0.077$\pm$0.000 & \textbf{0.202}$\pm$0.001 & \textbf{0.048}$\pm$0.000 & \textbf{0.075}$\pm$0.000 & \textbf{0.026}$\pm$0.001 & \textbf{0.104$\pm$0.001$^*$} & \textbf{0.019}$\pm$0.000 & \textbf{0.037$\pm$0.000$^*$} \\
\bottomrule
\end{tabular}}
\vspace{-0.2cm}
% \vspace{-5pt}
\end{table*}

\subsubsection{Image Classification}
\begin{table*}[t]
\centering
\caption{Overall performance comparison on Image Classication task. The highest accuracies are in Boldface.}
\vspace{-0.3cm}
\label{tab:overall_performance_on_image_classification}
% \resizebox{\textwidth}{!}{%
\begin{tabular}{c|ccccc}
\toprule
\textbf{CIFAR10} & 10\% & 20\% & 30\% & 40\% & 50\% \\
\midrule
Normal &89.01$\pm$0.41\% &87.44$\pm$0.85\% &86.06$\pm$0.55\% &84.15$\pm$0.47\% &81.83$\pm$0.77\% \\
MW-Net & 89.33$\pm$0.28\% &88.36$\pm$0.48\% &86.70$\pm$0.39\% &85.01$\pm$0.12\% &82.87$\pm$0.64\% \\
ITLM & 89.45$\pm$0.36\% &87.79$\pm$0.13\% &86.22$\pm$0.46\% &84.95$\pm$0.11\% &82.12$\pm$0.34\% \\
DeCA(p) & \textbf{90.32$\pm$0.18\%} &\textbf{88.78$\pm$0.20\%} &\textbf{87.07$\pm$0.19\%} &\textbf{86.62$\pm$0.26\%} &\textbf{84.32$\pm$0.32\%} \\
\bottomrule
\toprule
\textbf{FMNIST}& 10\% & 20\% & 30\% & 40\% & 50\% \\
\midrule
Normal & 87.01$\pm$0.02\% &86.42$\pm$0.03\% &85.58$\pm$0.02\% &84.89$\pm$0.02\% &83.62$\pm$0.02\% \\
MW-Net & 88.01$\pm$0.08\% &86.54$\pm$0.10\%&86.81$\pm$0.08\% &86.18$\pm$0.09\% &84.37$\pm$0.07\% \\
ITLM & 89.01$\pm$0.09\% &88.55$\pm$0.06\% &87.56$\pm$0.04\% &87.10$\pm$0.08\% &86.53$\pm$0.06\% \\
DeCA(p) & \textbf{89.47$\pm$0.05\%} &\textbf{89.44$\pm$0.03\%} &\textbf{88.86$\pm$0.04\%} &\textbf{88.72$\pm$0.06\%} &\textbf{88.25$\pm$0.07\%} \\
\bottomrule
\end{tabular}
% \vspace{-0.5cm}
% \vspace{-5pt}
\end{table*}

Table \ref{tab:overall_performance_on_image_classification} shows the results of the DeCA(p) and the baselines with the backbone ResNet-32 on CIFAR10 and Fashion-MNIST. The noises are added randomly. From the table, we can have following observations: (1) Our proposed method DeCA(p) consistently outperform the normal training setting and other denoising methods. (2) As the noise ratio gets larger, our methods get less affected by the noise ratio, which means the performance gain gets more significant. 

We also analyse how DeCA(p) affect the memorization of noisy samples, i.e. enhance the robustness of the mdoel. Figure \ref{fig:Analysis_on_memorizing_noisy_samples} shows the learning curve of the baselines and the proposed DeCA(p) with GMF as the target model on Modcloth, Adressa and Electronics for recommendation, and with ResNet32 on CIFAR10 for image classification. From the results, we could find that the problem of memorizing noisy samples is severe in normal training and also in other denoising methods. However, for DeCA(p), the performance remain stable and better along the whole training stage. The results demonstrate that the proposed DeCA(p) successfully prevent the model from being affected by noisy samples. Besides, even though the other methods may achieve the accuracy that is compatible to ours, the validation set may still fail to help choose the checkpoint with the highest accuracy. Thus our method DeCA(p) is still needed. 
\begin{figure}[t]
% \vspace{-0.8cm}
\centering     %%% not \center
\subfigure[GMF on Modcloth]{\label{fig:GMF_modcloth_recall}\includegraphics[width=0.24 \linewidth]{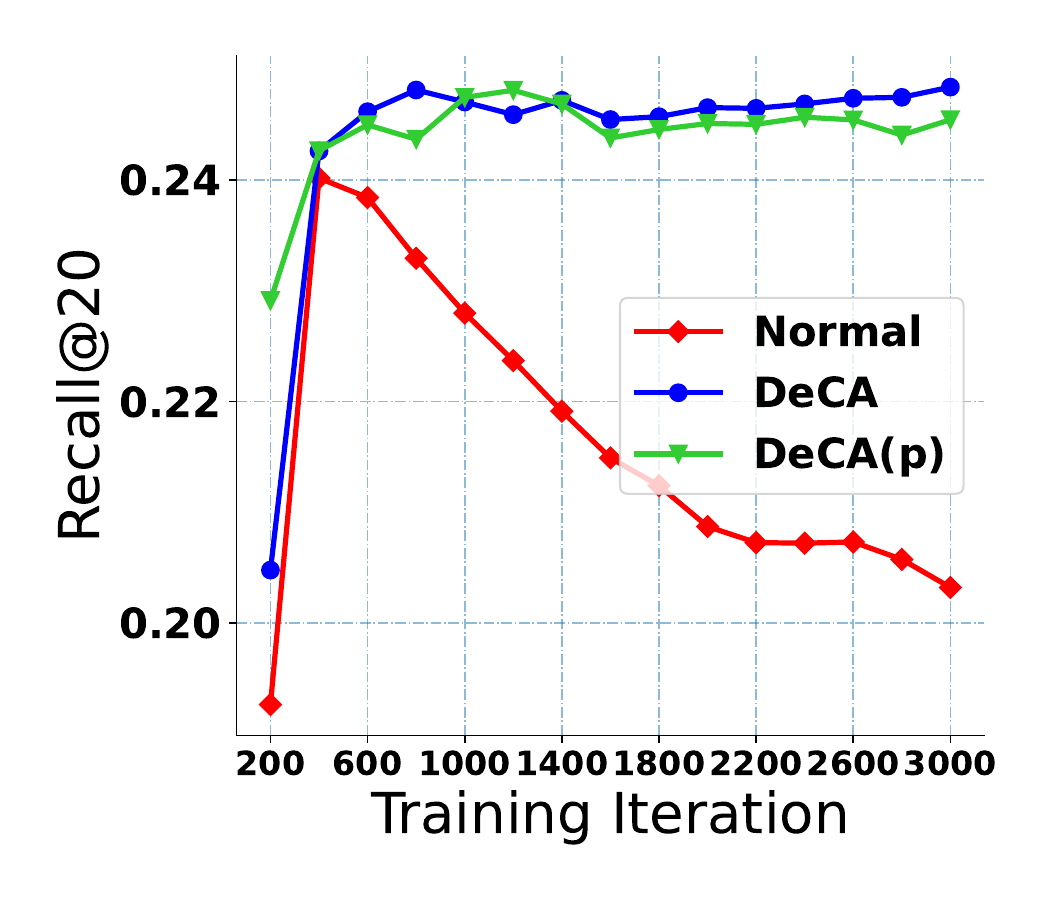}}
\subfigure[Adressa]{\label{fig:GMF_adressa_recall}\includegraphics[width=0.24\linewidth]{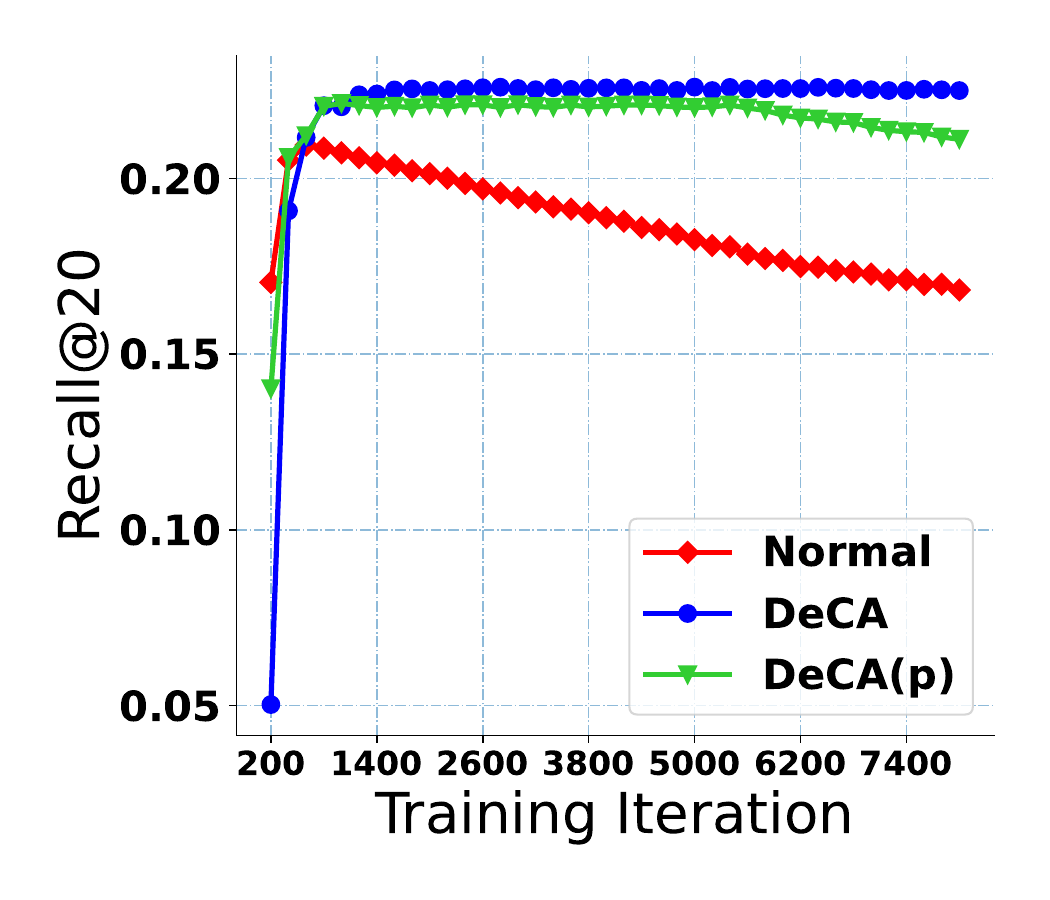}}
\subfigure[Electronics]{\label{fig:GMF_electronics_recall}\includegraphics[width=0.24\linewidth]{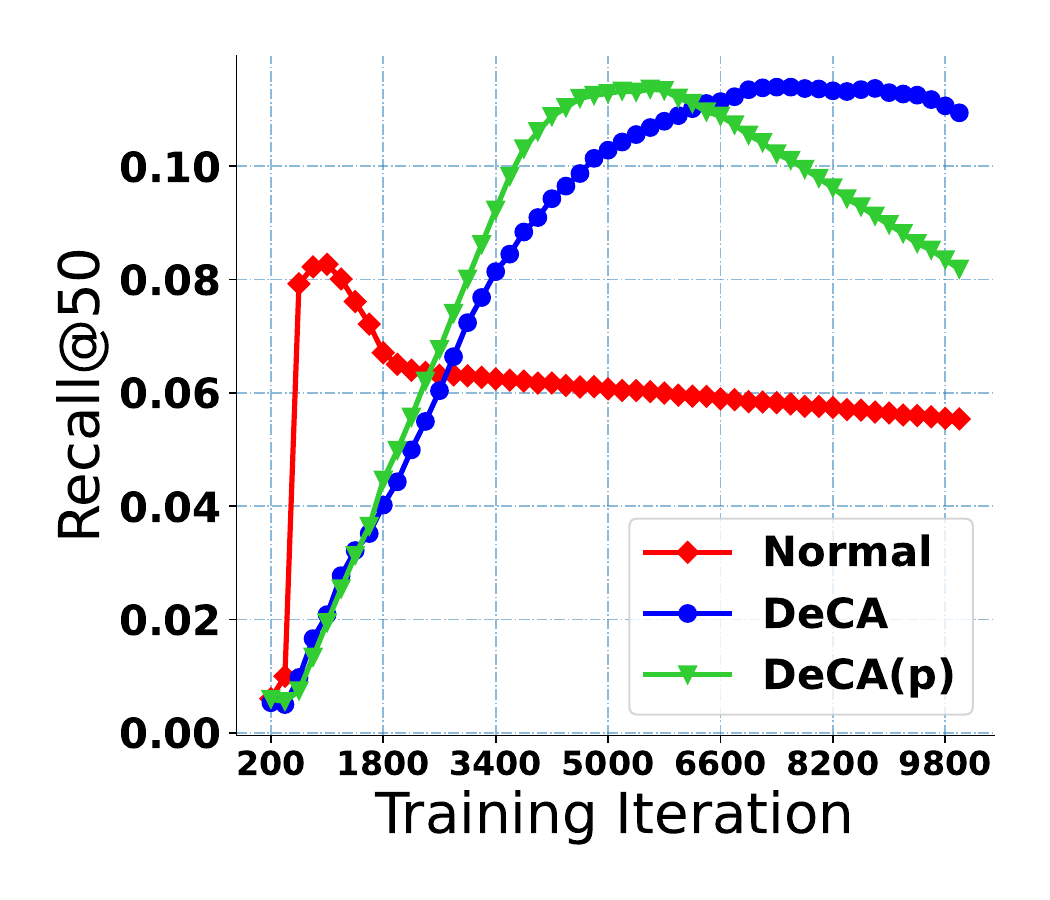}}
\subfigure[ResNet on CIFAR10]{\label{fig:CIFAR10_acc}\includegraphics[width=0.24 \linewidth]{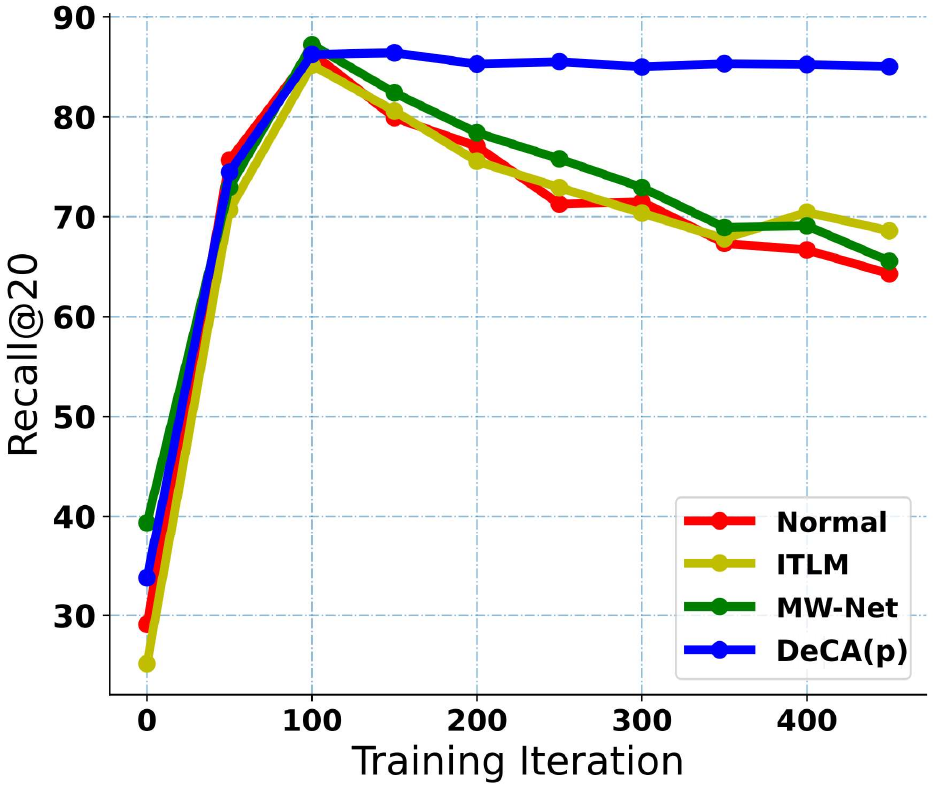}}
% \includegraphics[width=\linewidth]{figures/GMF_modcloth.png}
% \subfigure[Adressa]{\label{fig:GMF_adressa_recall}\includegraphics[width=0.494\linewidth]{pictures/GMF_adressa_recall.png}}
\vspace{-0.2cm}
\caption{Model performance along the training process.}
% \vspace{-0.5cm}
\label{fig:Analysis_on_memorizing_noisy_samples}
\end{figure}

% Besides, we also analyse how DeCA(p) affect the memorization of noisy samples. Figure \ref{fig:Analysis_on_memorizing_noisy_samples} shows the learning curve of normal training and the proposed DeCA and DeCA(p) on Modcloth, Adressa and Electronics when using GMF as the target recommender. We can see that in normal training, performance on the clean test set decreases in the late training stage. The reason is as the training goes on, the model tends to memorize all samples in the dataset, both noisy and clean ones. However, we can see that both DeCA and DeCA(p) remain more stable and better performance along the whole training stage. The results demonstrate that the proposed DeCA and DeCA(p) successfully prevent the model from being affected by noisy samples. 

% \begin{figure}
% \centering     %%% not \center
% \subfigure[Modcloth]{\label{fig:GMF_modcloth_recall}\includegraphics[width=0.32\linewidth]{figures/GMF_modcloth_recall_new.pdf}}
% \subfigure[Adressa]{\label{fig:GMF_adressa_recall}\includegraphics[width=0.32\linewidth]{figures/GMF_adressa_recall_new.pdf}}
% \subfigure[Electronics]{\label{fig:GMF_electronics_recall}\includegraphics[width=0.32\linewidth]{figures/GMF_electronics_recall_new.pdf}}
% \vspace{-0.4cm}
% \caption{Recall along the training process}
% \vspace{-15pt}
% \label{fig:Analysis_on_memorizing_noisy_samples}
% \end{figure}

\subsection{Model Investigation (RQ2)}
\subsubsection{Ablation Study} % (fold)
\label{ssub:ablation_study}
\begin{table*}[t]
\centering
\caption{Performance comparison when considering one sub-task.
DeCA-DP and DeCA-DN denote training DeCA with only either DP or DN. DeCA(p)-DP and DeCA(p)-DN denote training DeCA(p) with only either DP or DN.}
\vspace{-0.3cm}
\label{tab:Ablation_study_with_all_models_on_modcloth}
\resizebox{\textwidth}{!}{%
\begin{tabular}{c|cccc|cccc|cccc|cccc}
  \hline
& \multicolumn{4}{c|}{GMF} & \multicolumn{4}{c|}{NeuMF} & \multicolumn{4}{c|}{CDAE} & \multicolumn{4}{c}{LightGCN} \\ 
& R@5 & R@20 &N@5 &N@20 & R@5 & R@20 &N@5 &N@20 & R@5 & R@20 &N@5 &N@20 & R@5 & R@20 &N@5 &N@20 \\
\hline
\hline
DeCA-DP & 0.072 & 0.245 & 0.050 & 0.098 & 0.089 & 0.256 & 0.061 & 0.109 & 0.077 & 0.236 & 0.050 & 0.095 & 0.066 & 0.223 & 0.042 & 0.086 \\
DeCA-DN & 0.051 & 0.187 & 0.032 & 0.071 & 0.045 & 0.198 & 0.027 & 0.070 & 0.066 & 0.236 & 0.044 & 0.093 & \textbf{0.068} & \textbf{0.240} & \textbf{0.047} & \textbf{0.096} \\
DeCA & \textbf{0.072} & \textbf{0.245} & \textbf{0.050} & \textbf{0.099} & \textbf{0.099} & \textbf{0.268} & \textbf{0.065} & \textbf{0.113} & \textbf{0.086} & \textbf{0.250} & \textbf{0.056} & \textbf{0.102} & 0.064 & 0.221 & 0.041 & 0.085 \\
\hline
DeCA(p)-DP & \textbf{0.075} & 0.243 & 0.052 & 0.099 & \textbf{0.094} & \textbf{0.268} & \textbf{0.064} & \textbf{0.113} & \textbf{0.091} & 0.245 & \textbf{0.059} & 0.102 & 0.065 & 0.220 & 0.042 & 0.086 \\
DeCA(p)-DN & 0.060 & 0.221 & 0.039 & 0.084 & 0.092 & 0.255 & 0.062 & 0.108 & 0.073 & 0.238 & 0.050 & 0.097 & 0.072 & \textbf{0.240} & 0.049 & 0.096 \\
DeCA(p) & 0.074 & \textbf{0.247} & \textbf{0.052} & \textbf{0.100} & 0.087 & 0.265 & 0.059 & 0.110 & 0.089 & \textbf{0.251} & 0.057 & \textbf{0.103} & \textbf{0.073} & 0.235 & \textbf{0.051} & \textbf{0.096} \\
\hline
\end{tabular}}
\end{table*}
For recomendation tasks, DeCA and DeCA(p) utilize an iterative training routine that contains two sub-tasks DP and DN. In this part, we discuss how these two sub-tasks contribute to the whole framework and how they perform separately. In this part, we only pay attention to the recommendation tasks since Denoising Positive and Denoising Negative have their specific meanings in recommmendation. DP means the user click the item for some reason rather than the user likes the item; DN means the user does not click the item possibly due to the exposure bias. However, in image classification, the noises in different classes are equivalent in our setting, thus we do not conduct the ablations on image classification setting.
The results on Modcloth dataset are shown in Table \ref{tab:Ablation_study_with_all_models_on_modcloth}. 
We can have following observations: 
\begin{itemize}[leftmargin=*]
\item In most cases, DeCA and DeCA(p) are better than only considering one sub-task of denoising. Some abnormal cases could be attributed to the instability of DN.
During each training epoch, the interacted samples are fixed while the sampled negative instances could change frequently.
% according to the negative sampling strategy
Since DN aims to denoise noisy negative examples from the sampled missing interactions, the performance of DN could be unstable. 
\item DP is better than DN in most cases, except for the LightGCN model. This observation to some extent indicates that finding noisy examples from interacted samples is much easier than finding potential positive preference from the massive uninteracted missing samples.
\end{itemize}

\subsubsection{Hyperparameter Study.} % (fold)
\begin{figure}
\centering     %%% not \center
\subfigure[DeCA - Adressa]{\label{fig:hyper_GMF_adressa_DeCA}\includegraphics[width=0.240\linewidth]{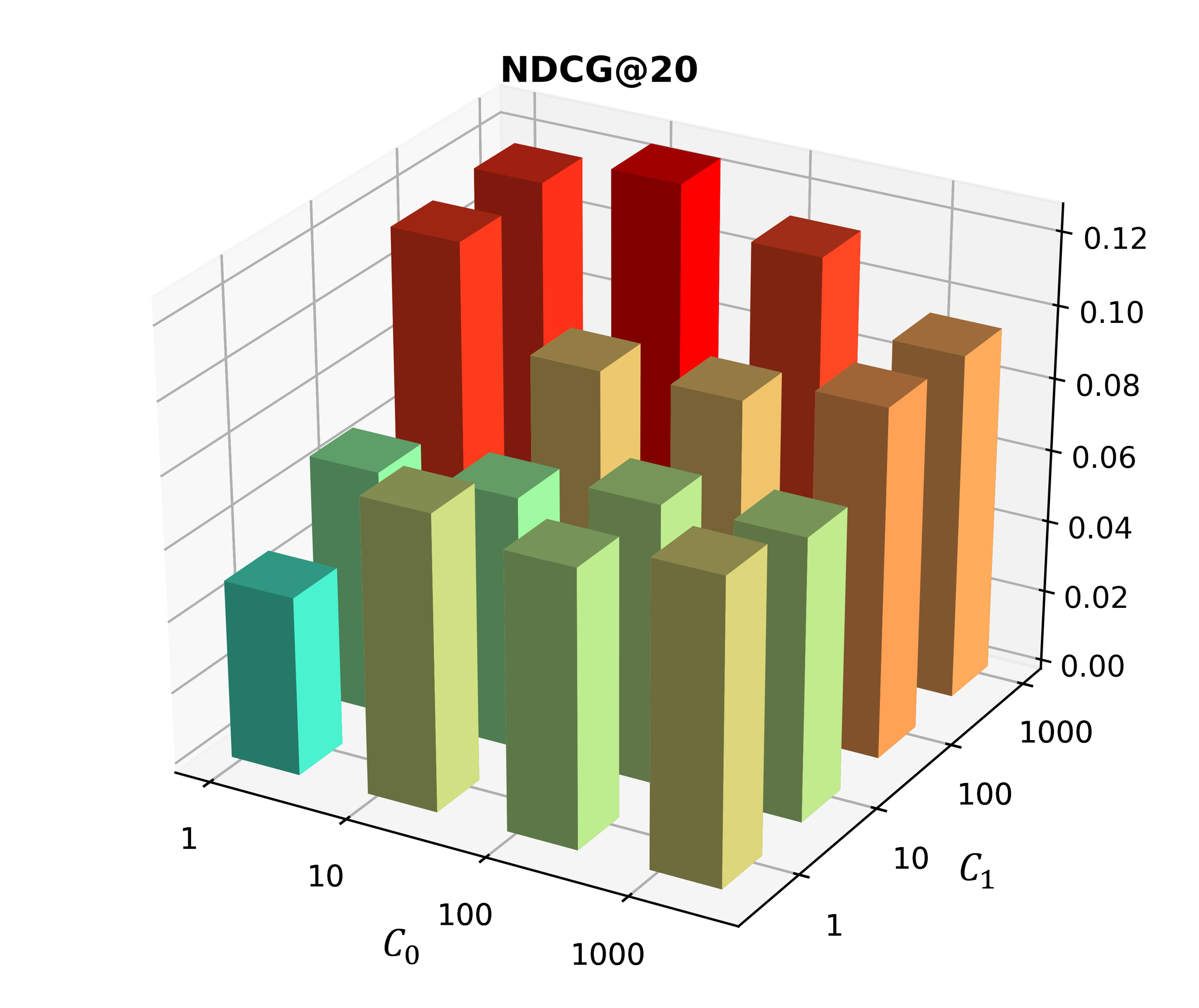}}
% \hfill
\subfigure[DeCA(p) - Adressa]{\label{fig:hyper_GMF_adressa_DeCA(p)}\includegraphics[width=0.240\linewidth]{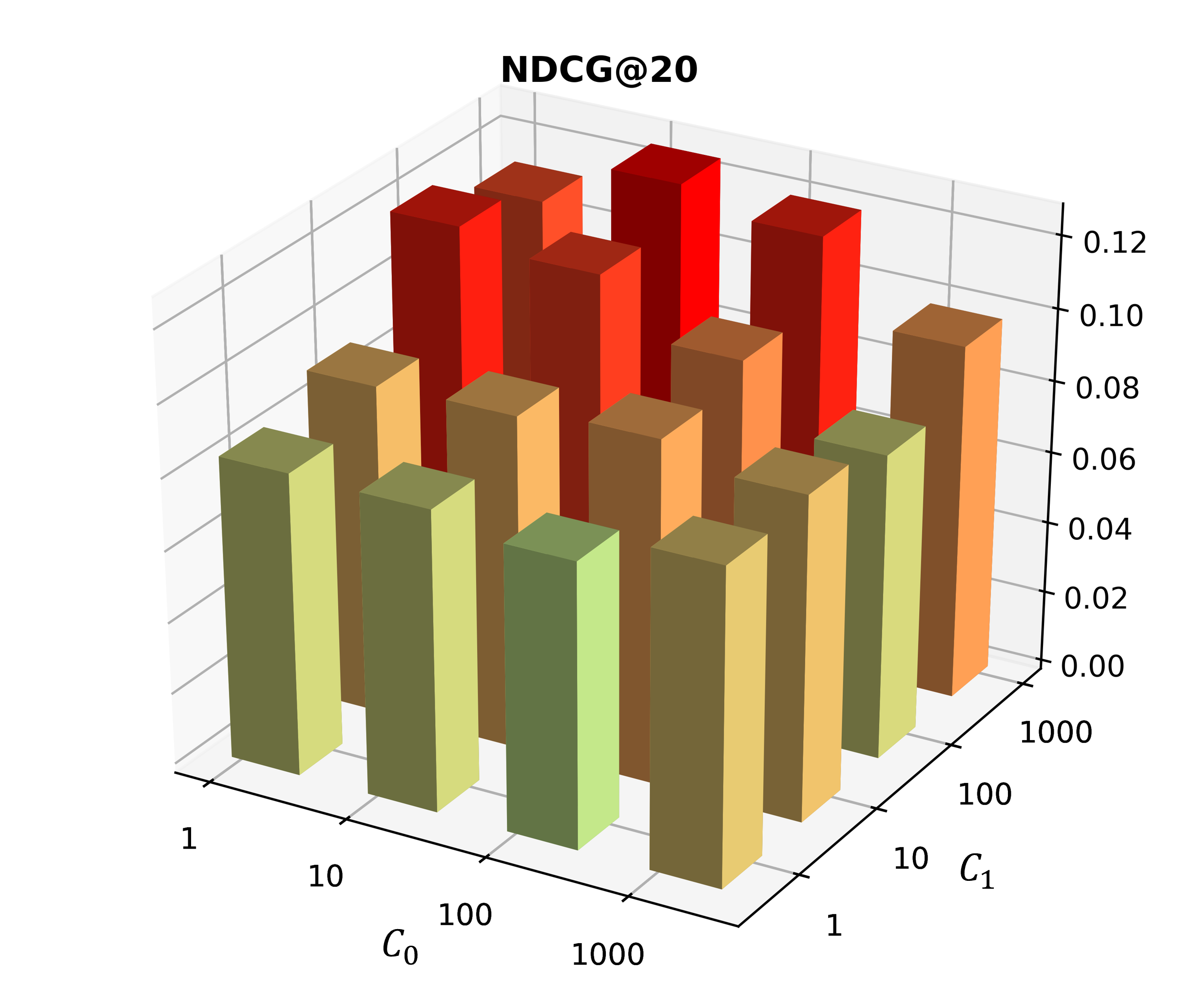}}
\hfill
\subfigure[DeCA - MovieLens]{\label{fig:DPI_movielens_hyer}\includegraphics[width=0.220\linewidth]{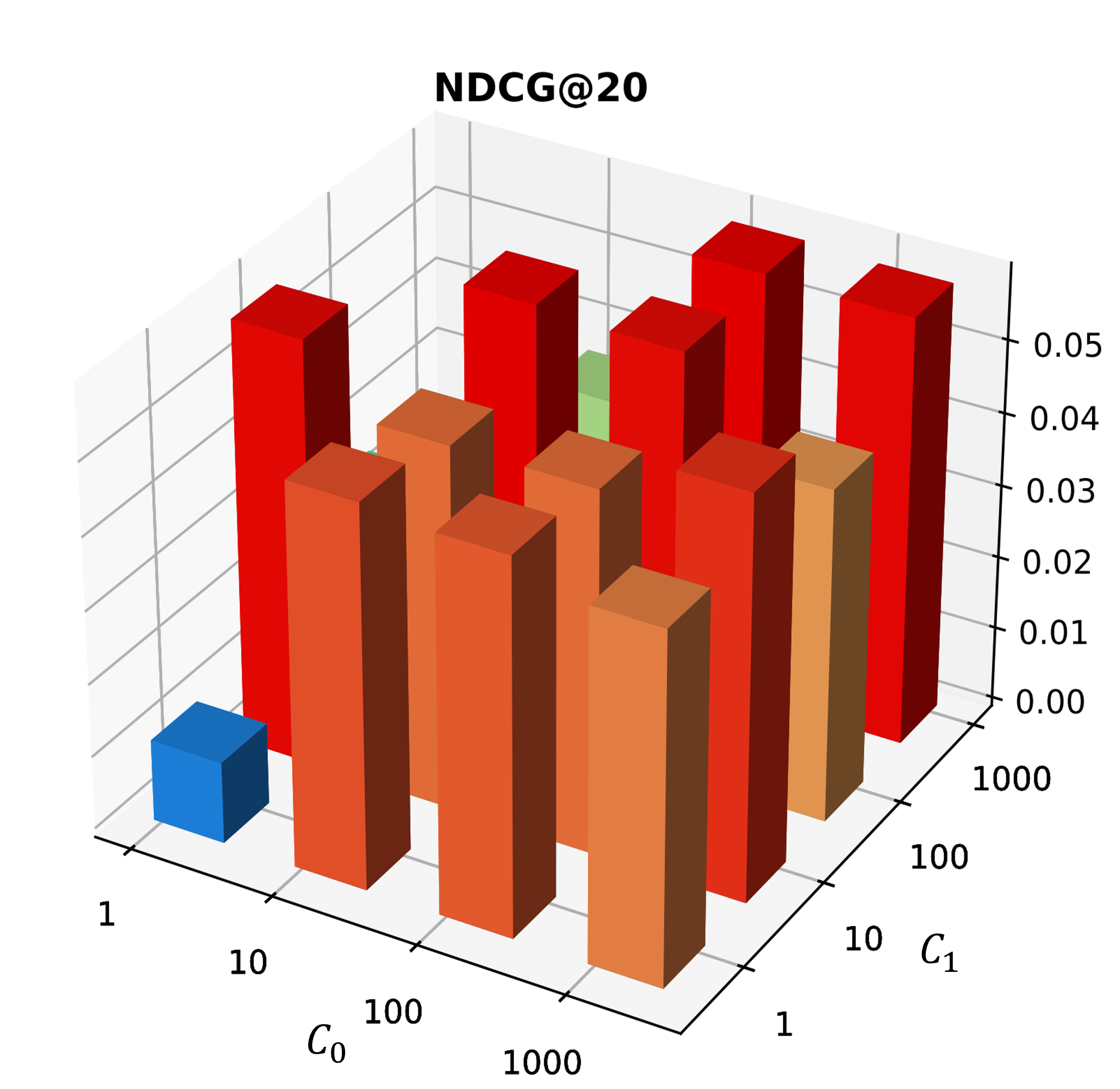}}
\hfill
\subfigure[DeCA(p) - MovieLens]{\label{fig:DVAE_movielens_hyer}\includegraphics[width=0.220\linewidth]{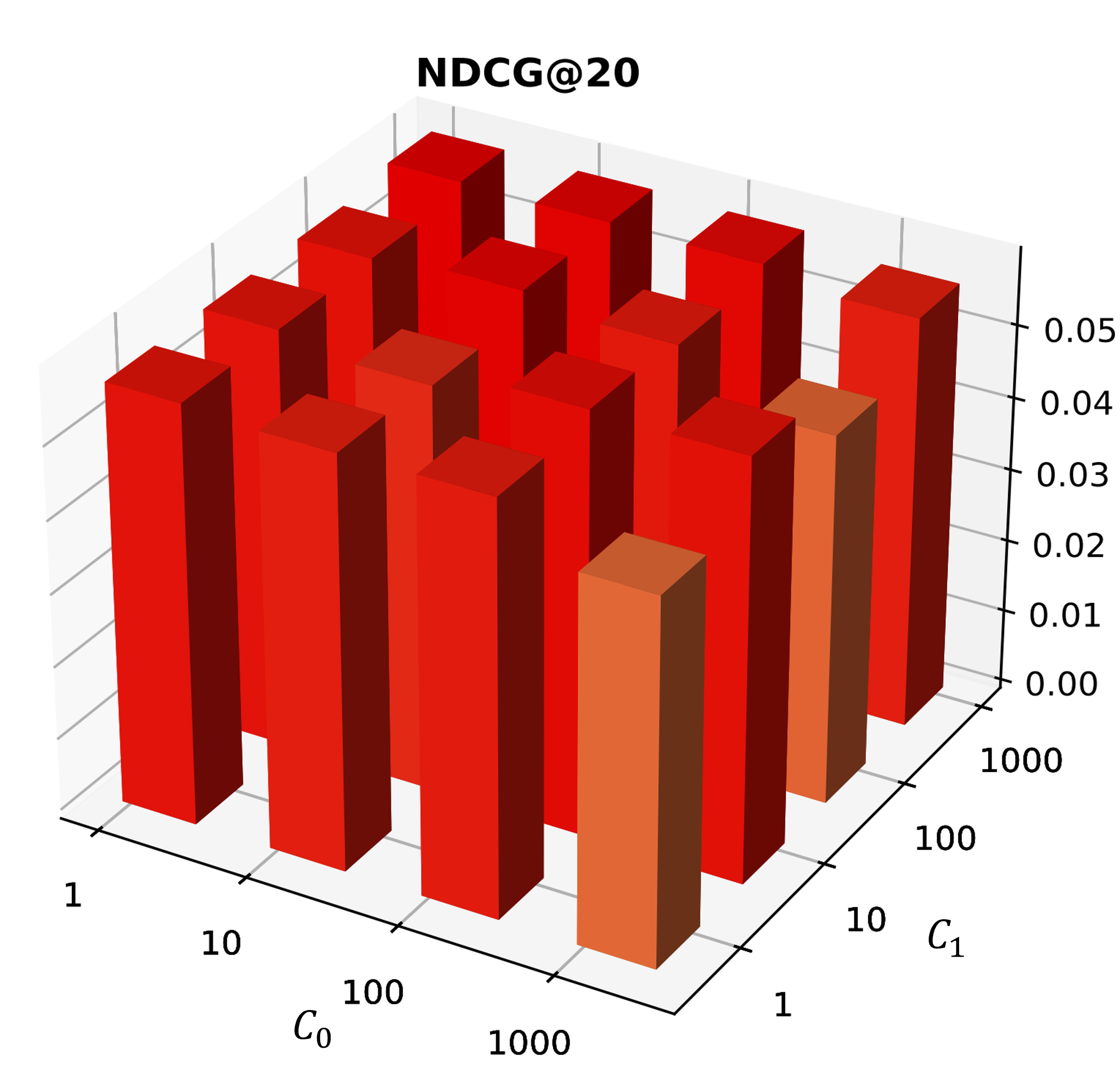}}
\subfigure[DeCA - Modcloth]{\label{fig:DPI_GMF_modcloth_hyper}\includegraphics[width=0.240\linewidth]{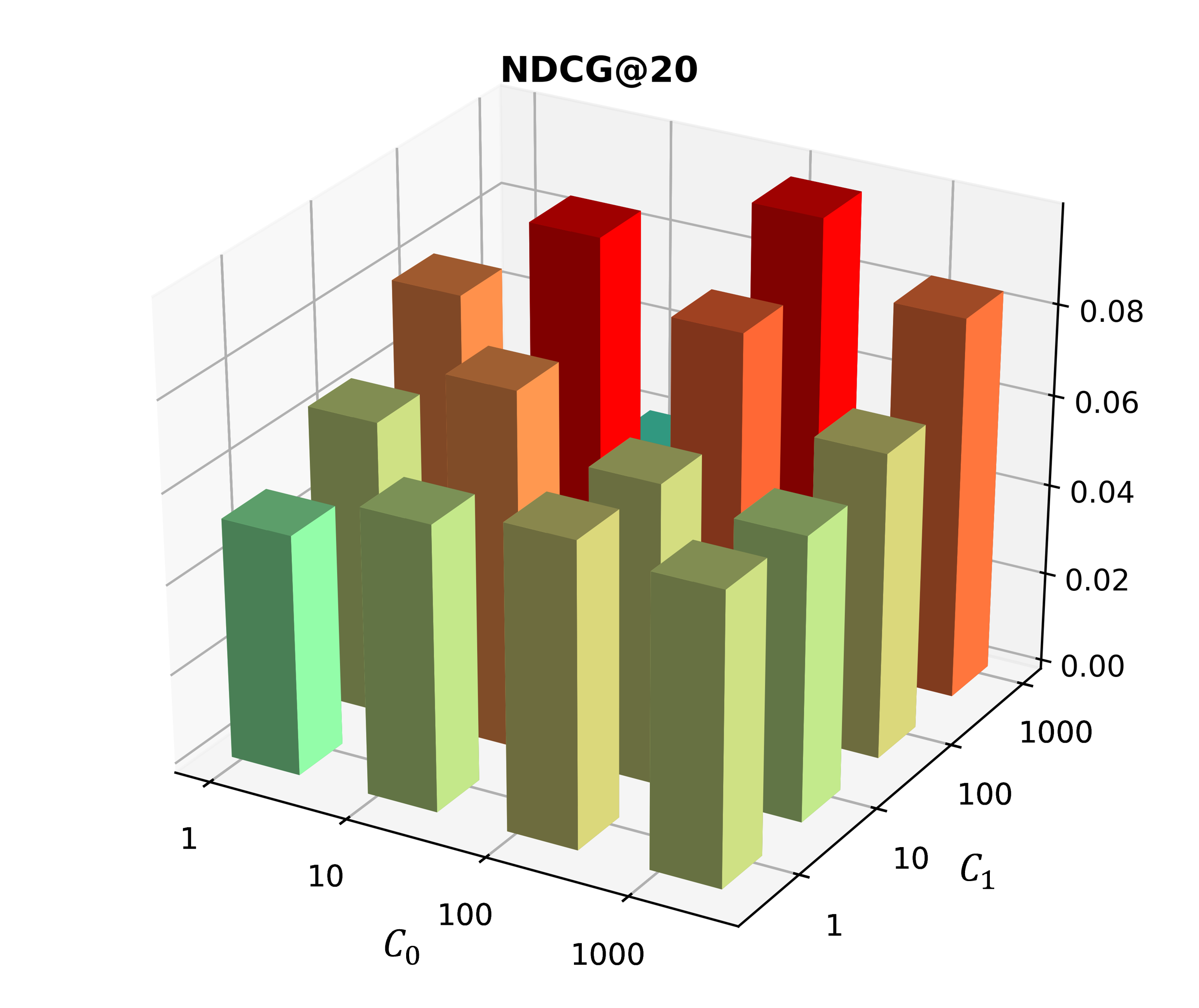}}
\subfigure[DeCA(p) - Modcloth]{\label{fig:DVAE_GMF_modcloth_hyper}\includegraphics[width=0.240\linewidth]{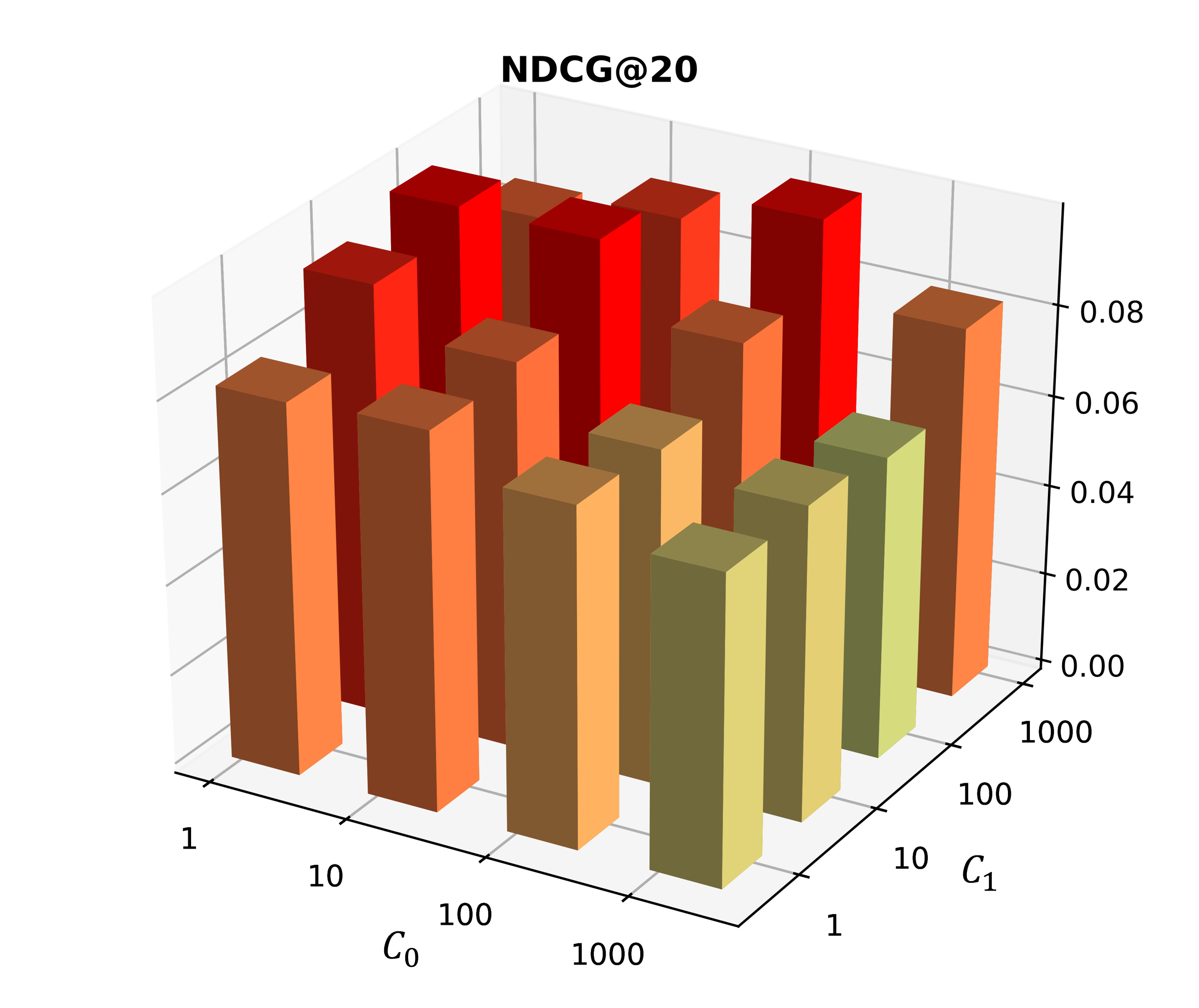}}
\subfigure[DeCA - Electronics]{\label{fig:DPI_GMF_electronics_hyper}\includegraphics[width=0.240\linewidth]{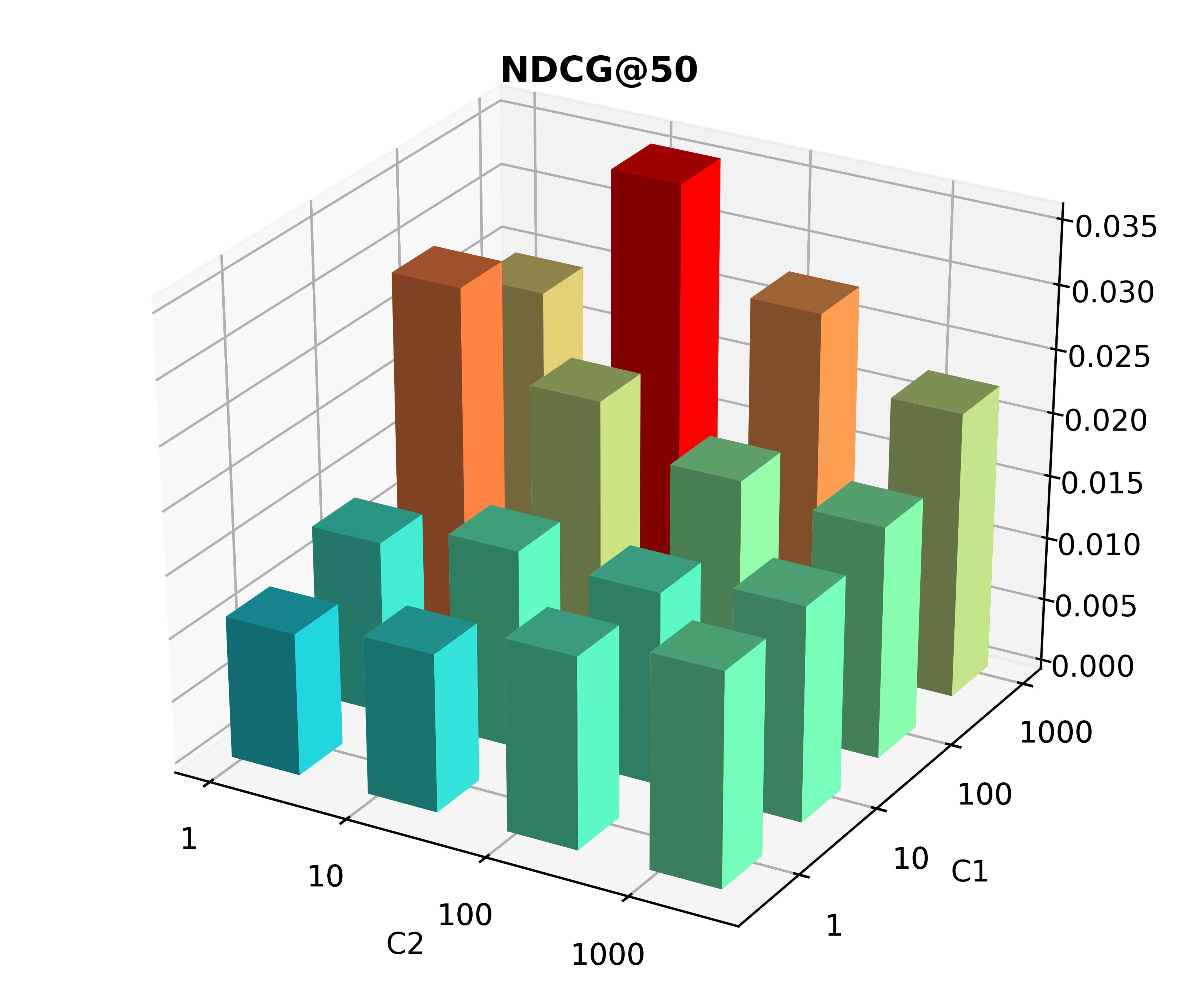}}
\subfigure[DeCA(p) - Electronics]{\label{fig:DVAE_GMF_electronics_hyper}\includegraphics[width=0.240\linewidth]{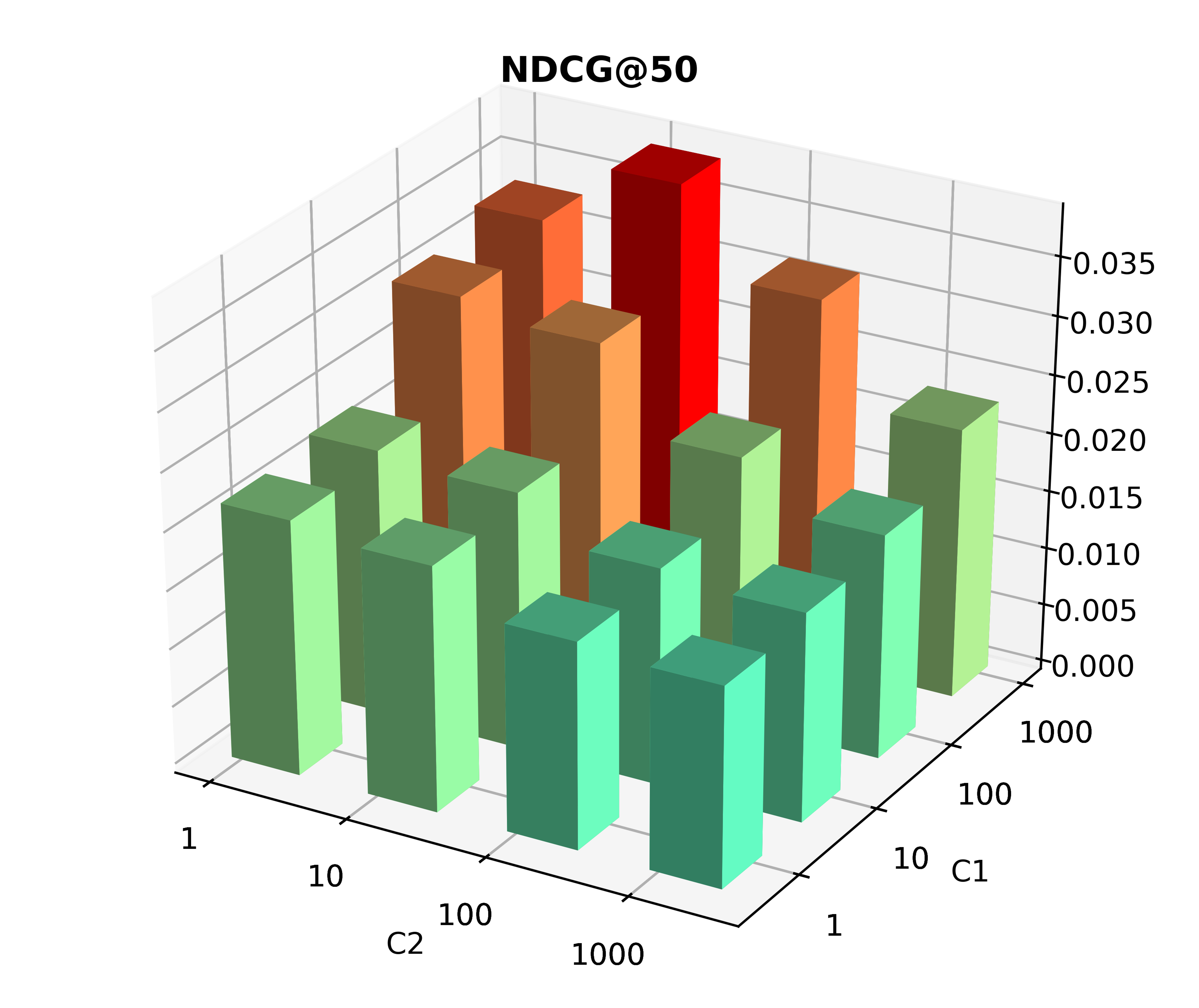}}
\vspace{-0.3cm}
\caption{Hyperparameter study on Adressa, MovieLens, Modcloth and Electronics.}
\vspace{-0.3cm}
\label{fig:Hyper_parameter_sensitivity}
\end{figure}

\label{ssub:hyper_parameter_sensitivity}
% \begin{wrapfigure}{r}{0.6\textwidth}
\begin{figure}
% \vspace{-0.8cm}
\centering     %%% not \center
\subfigure[ResNet on FMNIST]{\label{fig:GMF_modcloth_hyper}\includegraphics[width=0.3 \linewidth]{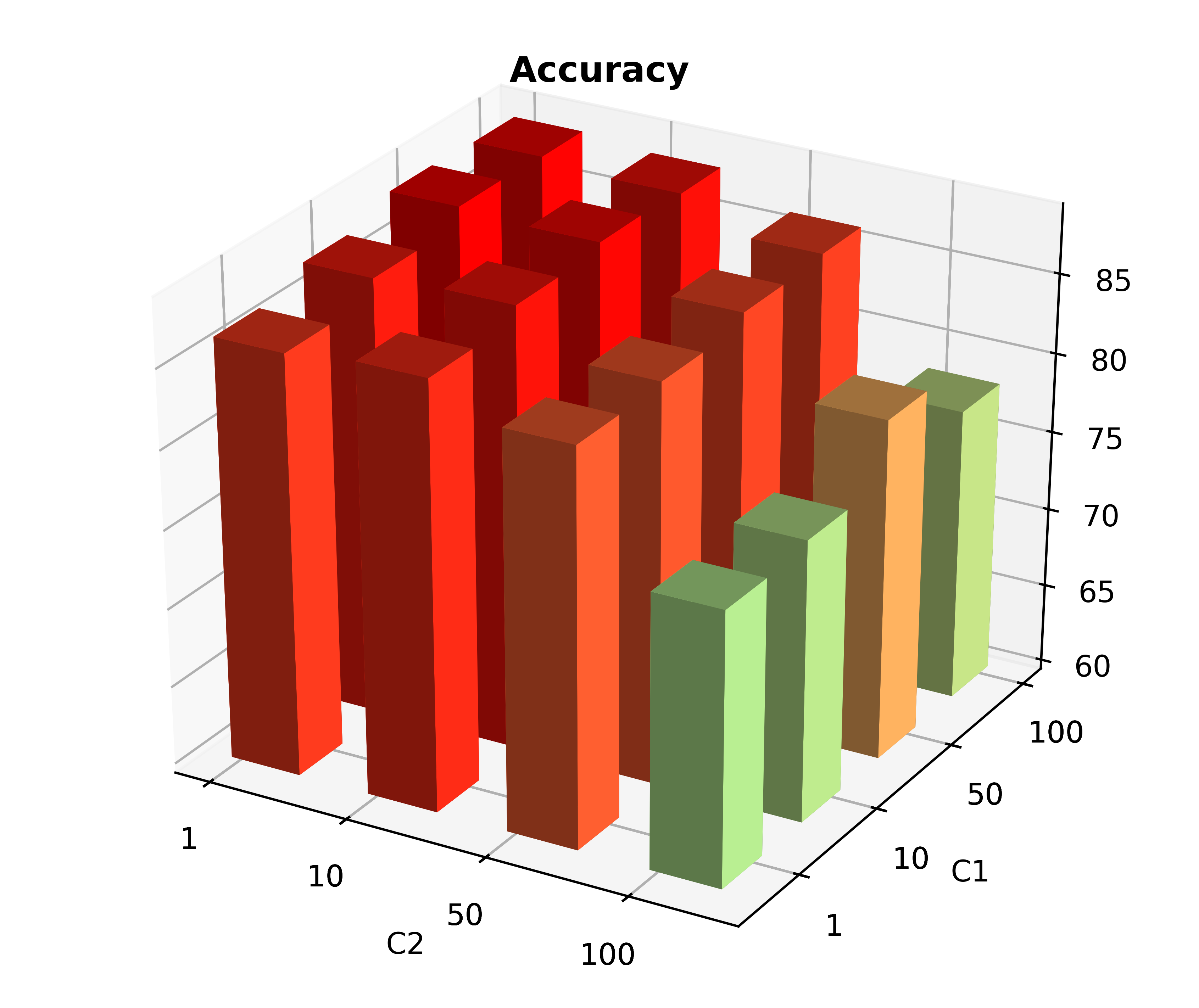}}
\subfigure[ResNet on CIFAR10]{\label{fig:CIFAR10_hyper}\includegraphics[width=0.3 \linewidth]{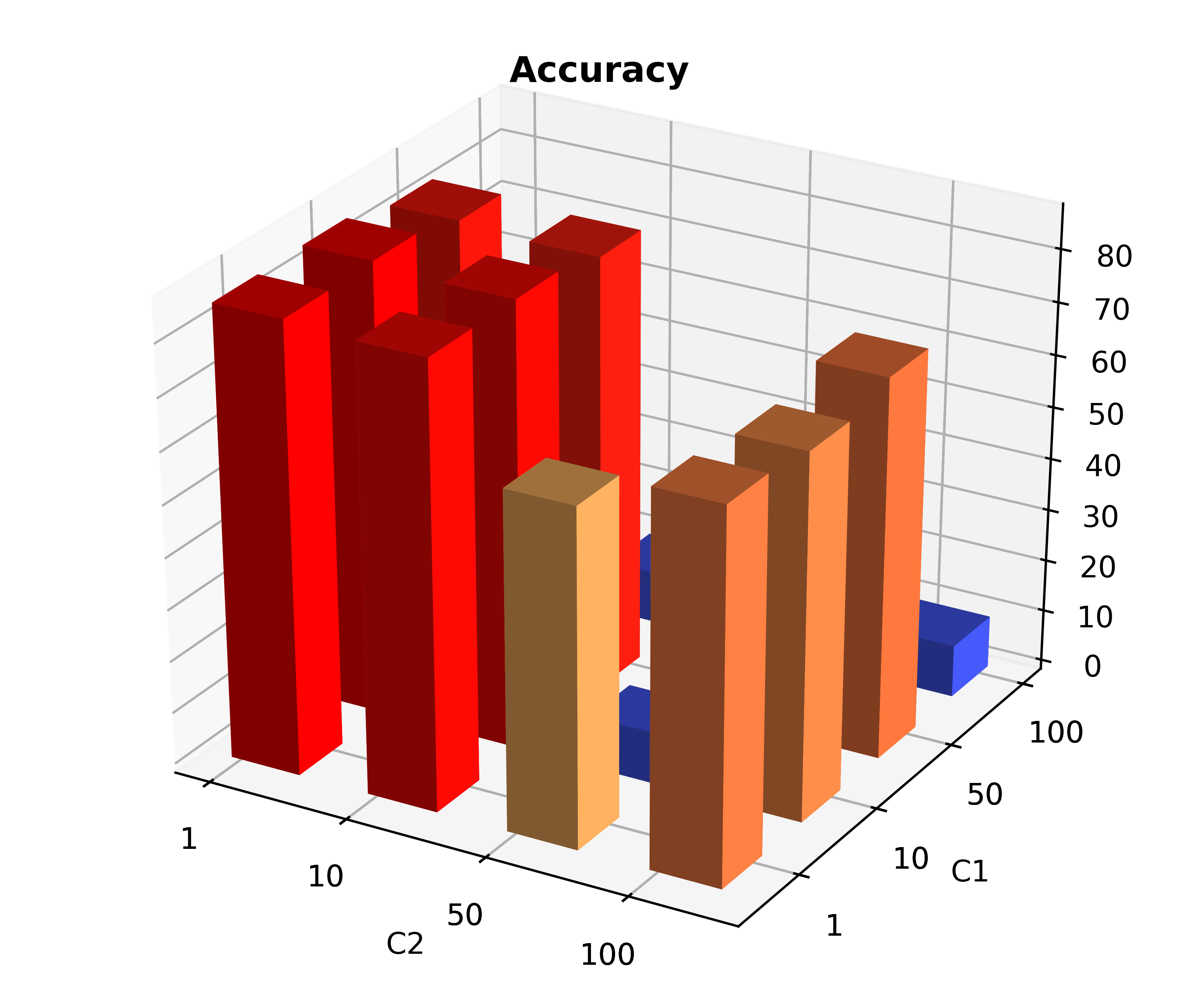}}
% \includegraphics[width=\linewidth]{figures/GMF_modcloth.png}
% \subfigure[Adressa]{\label{fig:GMF_adressa_recall}\includegraphics[width=0.494\linewidth]{pictures/GMF_adressa_recall.png}}
\vspace{-0.3cm}
\caption{Hyperparameter Sensitivity}
\vspace{-0.3cm}
% \vspace{-0.8cm}
\label{fig:Hyperparameter_sensitivity_on_classification}
\end{figure}
% \end{wrapfigure}
In this part, we conduct experiments to show the effect of $C_0$ and $C_1$, which are two hyper-parameters used in the iterative training routine.
More specifically, for recommendation, a larger $C_1$ means we are more confident that the negative samples are truly negative. A larger $C_0$ denotes that an interacted sample is more likely to be real positive. As for the classification, A larger $C_1$ means we are more confident that the examples of current class are clean, while a larger $C_0$ generally means we tend to believe that the examples from classes other than the current class are clean. Figure \ref{fig:Hyper_parameter_sensitivity} shows the results on all datasets when using GMF as the target recommendation model. We find that the optimal setting of $C_1$ (i.e., $C_1=1000$) is larger than the optimal setting of $C_0$ (i.e., $C_0=10 \text{ or } 100$). 
This observation indicates that the probability that a missing interaction is real negative is larger than the probability that an interacted sample is real positive. Figure \ref{fig:Hyperparameter_sensitivity_on_classification} shows the results of ResNet on MNIST and CIFAR10. \wyy{Note that we choose noise ratio to be $40\%$ when conducting the image classification experiments. As shown in the figure, Normally setting $C_1$ and $C_2$ to $1$ yields the best results. This could be because that $-\log P(\tilde{y}_i=k|y_i\neq k)$ and  $- \log P(\tilde{y}_i \neq k, \tilde{y}_i \neq y_i |y_i\neq k)$ are normally around $-\log (0.4) \approx 0.91$ since the noise ratio is 40\%. The experimental results also show that our results is consistent with the intuition.}

% \vspace{-10pt}
% subsubsection hyper_parameter_sensitivity (end)

% \begin{wraptable}{r}{0.6\textwidth}
\begin{table}
% \vspace{-0.5cm}
\centering
\caption{Effect of model selection on Modcloth.}
\label{tab:GMF_modcloth_gamma}
\begin{tabular}{c|c|cccc}
\hline
Method & $h$ and $h'$ & R@5 & R@20 & N@5 & N@20 \\
\hline 
\hline
\multicolumn{2}{c|}{Normal} & 0.0629 & 0.2246 &  0.0430 & 0.0884  \\
\hline
\multirow{3}{*}{DeCA}& MF&0.0717 & 0.2452 & 0.0500 & 0.0985 \\
& GMF& 0.0740 & 0.2453 & 0.0513 & 0.0989 \\
& NeuMF& 0.0747 & 0.2448 & 0.0519 & 0.0991 \\
\hline
\multirow{3}{*}{DeCA(p)}& MF& 0.0743 & 0.2465 & 0.0515 & 0.0996 \\
& GMF& 0.0740 & 0.2464 & 0.0508 & 0.0989 \\
& NeuMF& 0.0751 & 0.2464 & 0.0514 & 0.0992 \\
\hline
\end{tabular}
\end{table}
\subsubsection{Effect of Model Selection} % (fold)
\label{ssub:Effect_of_Model_Selection}
\wyy{In recommendation tasks}, we use MF as $h$ and $h'$ to model the probability $P(\tilde{\textbf{Y}}|\textbf{Y})$ as our default settings. Since the model capability of MF might be limited, in this part we conduct experiments to see how the performance would be if we use more complicated models for $h$ and $h'$. Table \ref{tab:GMF_modcloth_gamma} shows the result on Modcloth. Results on other datasets are provided in the supplement. The target recommendation model is GMF.  
We can see that replacing MF with more complicated models will not boost the performance significantly. The reason might be that modelling the probability $P(\tilde{\textbf{Y}} | \textbf{Y})$ is not a very complex task. Accomplishing this task with MF already works well. 
Besides, as we can see that no matter what model we use, the performance of the proposed DeCA and DeCA(p) are significantly better than normal training.

\begin{figure*}[h!]
\vspace{-0.2cm}
\subfigure[MovieLens-DeCA]{\label{fig:DPI_movielens_rating}\includegraphics[width=0.28\textwidth]{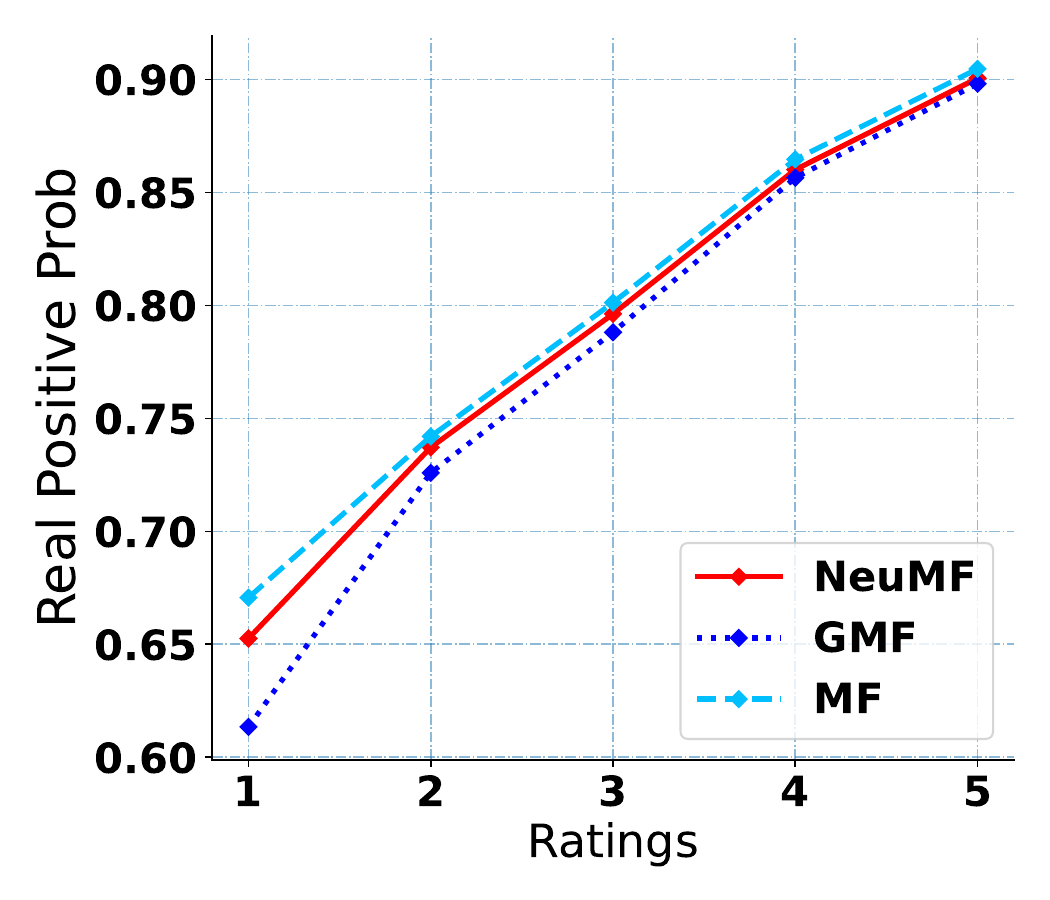}}
\subfigure[MovieLens-DeCA(p)]{\label{fig:DVAE_movielens_rating}\includegraphics[width=0.28\textwidth]{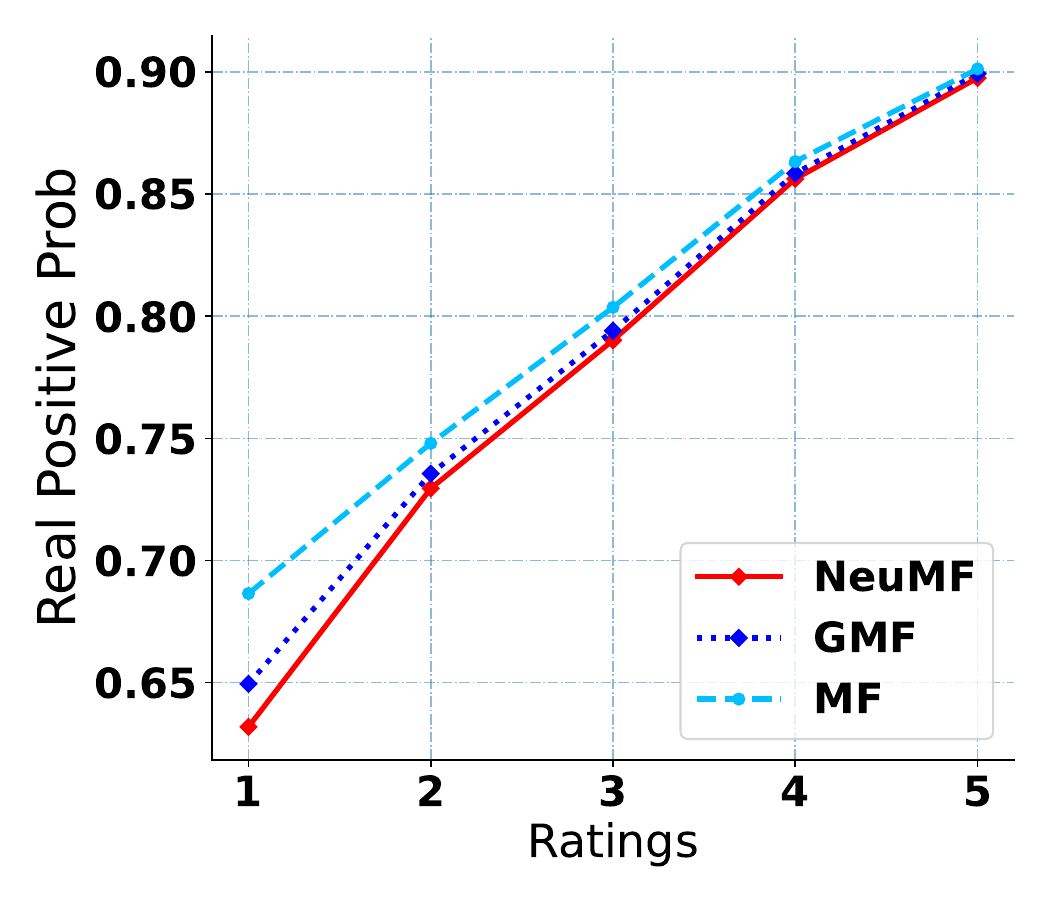}}
\subfigure[Modcloth-DeCA]{\label{fig:DPI_modcloth_rating}\includegraphics[width=0.28\textwidth]{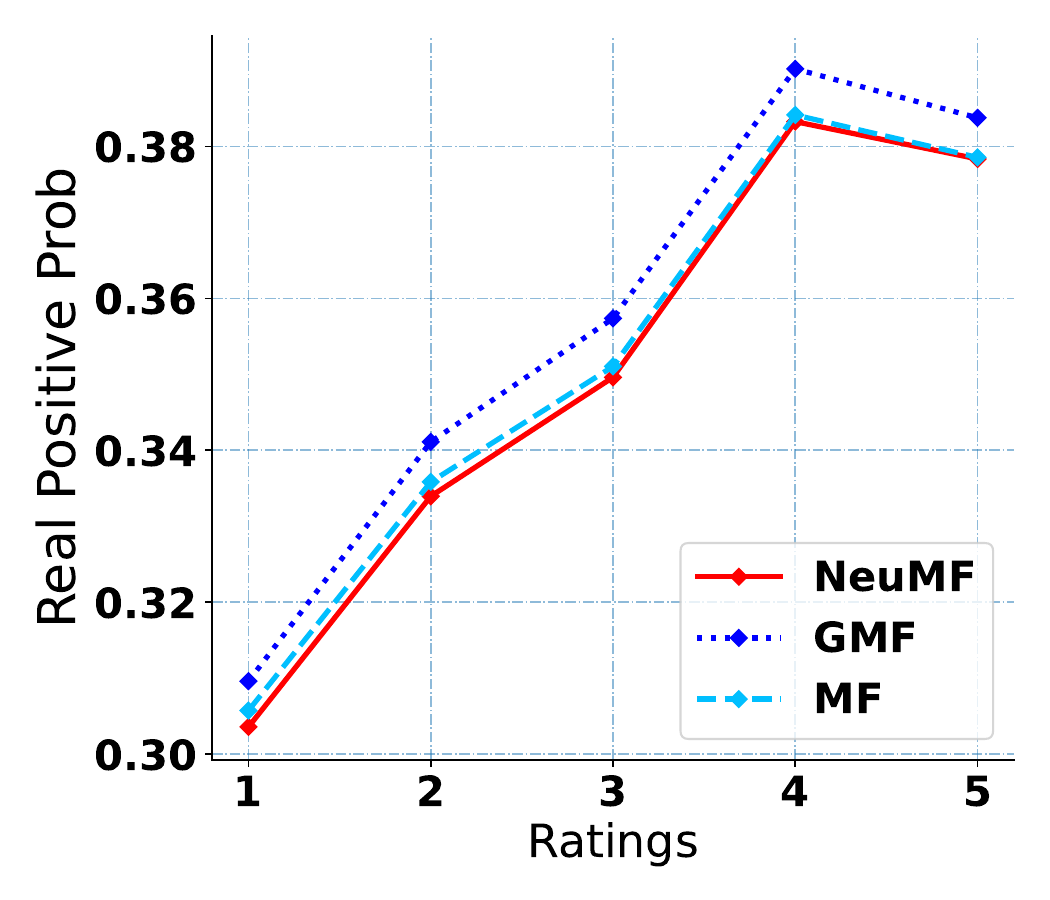}}
\subfigure[Modcloth-DeCA(p)]{\label{fig:DVAE_modcloth_rating}\includegraphics[width=0.28\textwidth]{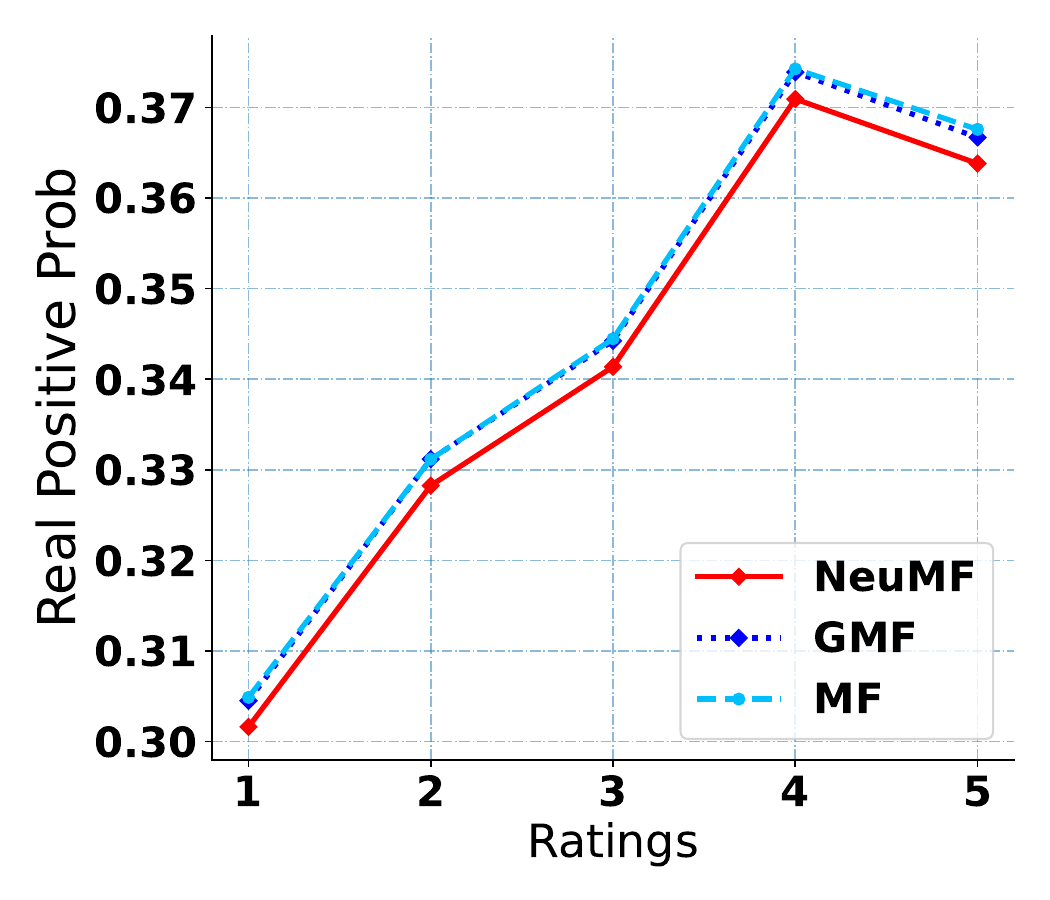}}
\subfigure[Electronics-DeCA]{\label{fig:DPI_electronics_rating}\includegraphics[width=0.28\textwidth]{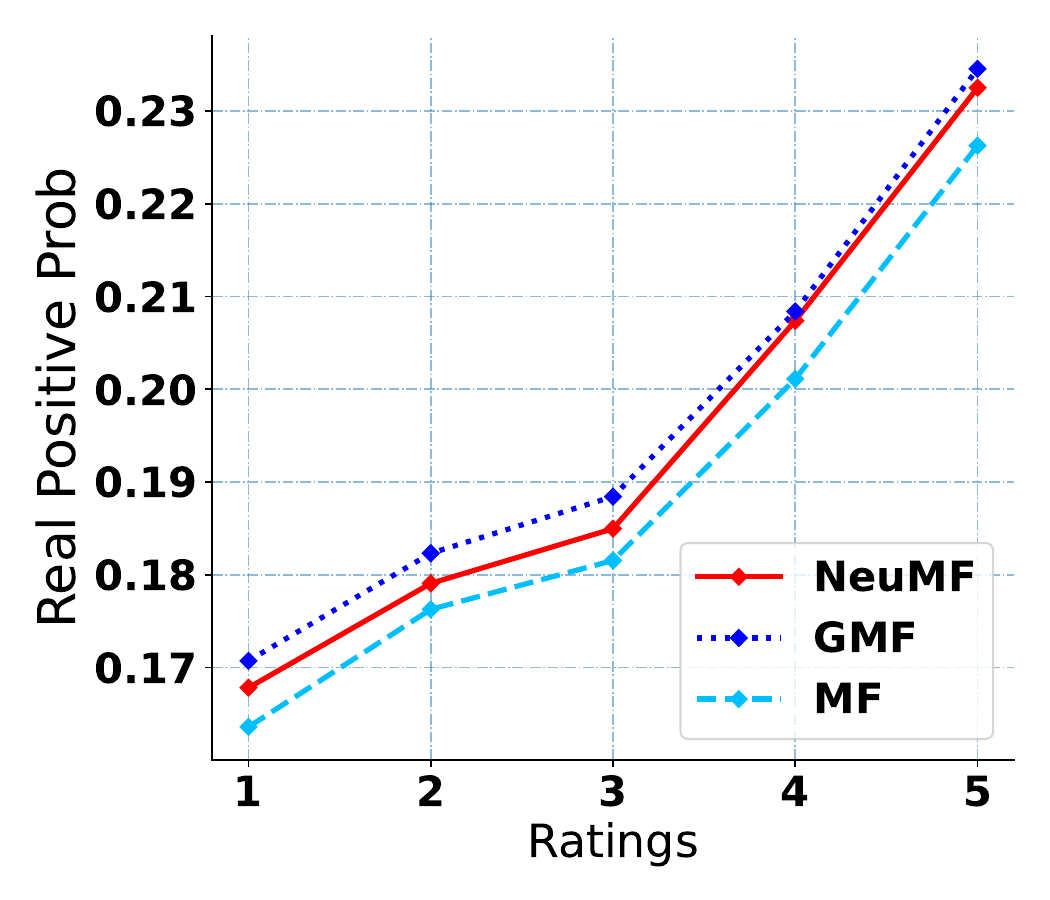}}
\subfigure[Electronics-DeCA(p)]{\label{fig:DVAE_electronics_rating}\includegraphics[width=0.28\textwidth]{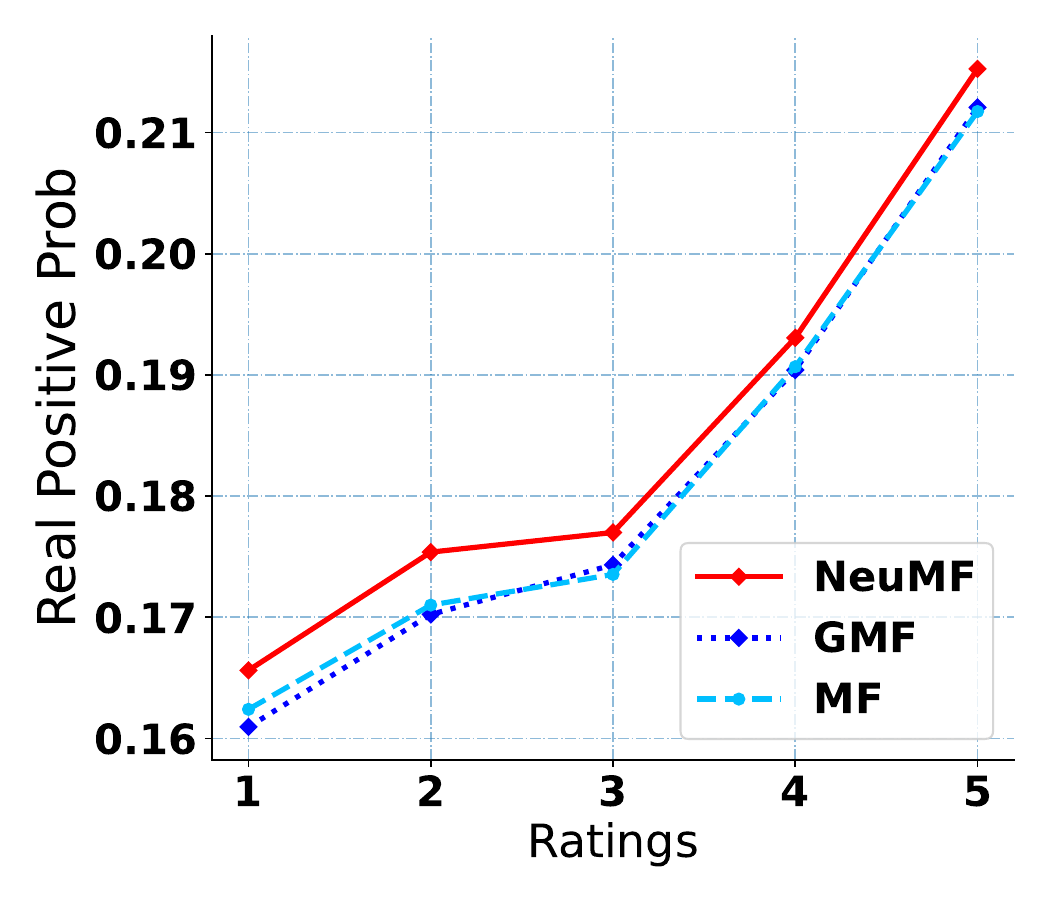}}
% \vspace{-5pt}
\caption{Mean real positive probability of different ratings on three datasets.}
% \vspace{-0.2cm}
\label{mean_real_positive_probability}
\end{figure*}

% subsubsection Effect_of_Model_Selection (end)
\vspace{-5pt}
\subsection{Method Interpretation (RQ3)} % (fold)
\label{ssub:method_interpretation}
In this part, we conduct experiments to see whether DeCA and DeCA(p) generate the reasonable user preference distribution given the corrupted data, which can be used to show the interpretability of our methods.
For DeCA, we have the following equation:
\begin{align*}
  P(y_{i}&=1|\tilde{y}_{i}=1) =\frac{P(\tilde{y}_{i}=1|y_{i}=1)P(y_{i}=1)}{P(\tilde{y}_{i}=1)}  \\
  =&\frac{h'_\psi(x_i) f_\theta(x_i)}{h'_\psi(x_i) f_\theta(x_i) + h_\phi(x_i)(1-f_\theta(x_i))}  \tag{\stepcounter{equation}\theequation} 
\end{align*}
where $x_i$ is the corresponding $i$-th user-item pair. For DeCA(p), $P(y_{i}=1|\tilde{y}_{i}=1)$ can be directly computed as $P(y_{i}=1|\tilde{y}_{i}=1)= f_\theta(x_i)$ with condition $\tilde{y}_{i}=1$.
$P(y_{i}=1|\tilde{y}_{i}=1)$ describes the probability of an interacted sample to be real positive. Figure \ref{mean_real_positive_probability} shows the relationship between ratings and this mean real positive probability on the datasets MovieLens, Modcloth and Electronics. 
We can see that the probability gets larger as ratings go higher, which is consistent with the impression that examples with higher ratings should be more likely to be real positive. 
%indicate that our methods are capable of detecting noisy samples, thus to avoid the model to be affected by these noisy samples. 
Besides, we can see that the probability trend is consistent with different $h_\phi$ and $h_\psi'$ models, which further demonstrate the generalization ability of our methods.

\section{Conclusion and Limitation}
\label{conclusion}
% In this paper, we propose probabilistic and variational denoising (PVD) for corrupted labels, which could be applied into classification problems. PVD is the extension of DeCA~\cite{DeCA} from binary classfication tasks in recommendation to multi-class classification. 
% The intuition behind PVD is that different models (different structures or same model with different random seeds) tend to make more consistent predictions on clean examples than noisy examples. 
% Based on this, we propose to denoise labels by minimizing the KL-divergence between the real label distributions parameterized by two models trained with different random seeds while maximizing the likelihood of data observation.
% The effectiveness of PVD is demonstrated on both binary classification tasks (recommendation systems) and multi-class classification tasks (image classification).
In this work, we propose model-agnostic training frameworks to learn robust machine learning models from corrupted labels. We find that different models tend to make more consistent agreement predictions for clean examples compared with noisy ones.
%have large variance in the predictions on noisy samples while make much more consistent predictions on clean ones. 
To this end, we propose \emph{denoising with cross-model agreement} (DeCA), which utilizes predictions from different models as the denoising signal. 
We start from the binary implicit feedback recommendation task and employ the proposed methods on four recommendation models and conduct extensive experiments on four datasets. 
Besides, we further extend the proposed method DeCA(p) for the multi-class scenario and conduct experiments on the image classification tasks with noisy labels. 
Experimental results demonstrate that our methods outperforms the previous baselines significantly. 
\wy{Regarding limitations, it is noteworthy that DeCA(p) training time is twice as long as regular training. Additionally, for classification problems with numerous classes, the process may become complex. Future work entails devising more efficient training methods for $h_\varphi$ to expedite the process.}

\clearpage

\bibliographystyle{ACM-Reference-Format}
\bibliography{sample-base}

\end{document}